\documentclass{article}
\usepackage{jfrExamplee}
\usepackage{graphicx}
\usepackage{apalike}
\usepackage{setspace}
\usepackage{graphics,epsfig,times,amsmath,amssymb,xcolor,multirow,amsmath,comment, algorithm, algorithmic,makecell}
\newcommand\keywords[1]{\textbf{Keywords}: #1}

\newcommand{\colo}{\color{black}}
\newcommand{\colt}{\color{black}}
\newcommand{\colm}{\color{black}}
\newcommand{\colb}{\color{black}}
\newcommand{\colc}{\color{black}}

\newcommand{\colr}{\color{black}}

\title{An Aerial Transport System in Marine GNSS-Denied Environment}

\author{
Jianjun Sun \\
School of Aerospace Engineering\\
Beijing Institute of Technology\\
Beijing 100081, China \\
\texttt{jianjun.sun@bit.edu.cn} \\
\And
Zhenwei Niu \\
Centre for Autonomous Robotic Systems\\
Khalifa University\\
Abu Dhabi 127788, UAE \\
\texttt{zhenwei.niu@ku.ac.ae} \\
\And
Yihao Dong \\
Centre for Autonomous Robotic Systems\\
Khalifa University\\
Abu Dhabi 127788, UAE\\
\texttt{yihao.dong@ku.ac.ae} \\
\And
Fenglin Zhang \\
School of Aerospace Engineering\\
Beijing Institute of Technology\\
Beijing 100081, China \\
\texttt{3220220075@bit.edu.cn} \\
\And
Muhayy Ud Din \\
Centre for Autonomous Robotic Systems\\
Khalifa University\\
Abu Dhabi 127788, UAE \\
\texttt{muhayyuddin.ahmed@ku.ac.ae} \\
\And
Lakmal Seneviratne \\
Centre for Autonomous Robotic Systems\\
Khalifa University\\
Abu Dhabi 127788, UAE \\
\texttt{lakmal.seneviratne@ku.ac.ae} \\
\And
Defu Lin \\
Centre for Autonomous Robotic Systems\\
School of Aerospace Engineering\\
Beijing Institute of Technology\\
Beijing 100081, China \\
\texttt{lindf@bit.edu.cn} \\
\And
Irfan Hussain\thanks{Corresponding author: Irfan Hussain.} \\
Centre for Autonomous Robotic Systems\\
Khalifa University\\
Abu Dhabi 127788, UAE \\
\texttt{irfan.hussain@ku.ac.ae} \\
\And
Shaoming He\thanks{Corresponding author: Shaoming He.} \\
School of Aerospace Engineering\\
Beijing Institute of Technology\\
Beijing 100081, China \\
\texttt{shaoming.he@bit.edu.cn} \\
}

\begin{document}

\maketitle
\keywords{Unmanned Aerial Vehicle, Vision-Based Landing, Localization, GNSS-Denied, Marine Transport, System Integration.}
\begin{abstract}
This paper presents an autonomous aerial system specifically engineered for operation in challenging marine GNSS-denied environments, aimed at transporting small cargo from a target vessel. In these environments, characterized by weakly textured sea surfaces with few feature points, {\colc chaotic deck oscillations} due to waves, and significant wind gusts, conventional navigation methods often prove inadequate.
Leveraging the DJI M300 platform, our system is designed to autonomously navigate and transport cargo while overcoming these environmental challenges.
{\colm In particular, this paper proposes an anchor-based localization method using ultrawideband (UWB) and QR codes facilities, which decouples the UAV's attitude from that of the moving landing platform, thus reducing control oscillations caused by platform movement. Additionally, a motor-driven attachment mechanism for cargo is designed, which enhances the UAV’s field of view during descent and ensures a reliable attachment to the cargo upon landing.
The system's reliability and effectiveness were progressively enhanced through multiple outdoor experimental iterations and were validated by the successful cargo transport during the {\colc 2024 Mohamed BinZayed International Robotics Challenge (MBZIRC2024)} competition.}
Crucially, the system addresses uncertainties and interferences inherent in maritime transportation missions without prior knowledge of cargo locations on the deck and with strict limitations on intervention throughout the transportation.
\end{abstract}

\section{Introduction}

With the rapid development of autonomous technology in recent years, the application of Unmanned Aerial Vehicles (UAVs) in cargo transport is expected to grow quickly \cite{ref:DAJ2020}. A significant portion of international cargo transport is marine-related, but the complex environment characterized by high winds, large waves, and other natural disasters often affects transport efficiency and the safety of human workers \cite{ref:YMI2020}. UAVs have great potential in this sector to undertake tasks that may be unsafe for humans or cause fatigue, owing to their increasing carrying capacity and high maneuverability \cite{ref:FUM2022}.

To achieve closed-loop feedback control for autonomous transport, UAVs need to accurately estimate their position and attitude relative to the surrounding environment \cite{ref:MAJ1997}. 
Currently, most platforms integrate position information from Global Navigation Satellite System (GNSS) receivers along with acceleration and angular rate measurements from IMUs \cite{ref:ABI2020}.
However, relying solely on GNSS for localization is not reliable. For example, {\colc GNSS signals can be obstructed by ship hulls, subject to electromagnetic interference, or even vulnerable to jamming and spoofing attacks}, leading to potential vulnerabilities in the aerial transport system \cite{ref:FAJ2018}.
These risks have led to the investigation of autonomous aerial systems designed for cargo transport in marine environments where access to GNSS signals is unavailable.
Developing an autonomous aerial system for marine GNSS-denied environments can significantly enhance the robustness and efficacy of cargo transport under challenging conditions.
By addressing the unique navigational constraints posed by such environments, these systems hold promise in mitigating the adverse effects of unpredictable weather patterns, turbulent sea conditions, and other environmental factors that traditionally impede maritime cargo operations.

Given these considerations, ASPIRE, the program management pillar of Abu Dhabi's Advanced Technology Research Council, hosts the Mohamed Bin Zayed International Robotics Challenge 2024 (MBZIRC2024) with the explicit objective of advancing research in autonomous robotic aerial vehicle technologies.
Compared to laboratory experiments, the assessment of robotic algorithms in this context proved to be considerably more rigorous. Each team was limited in the number of trials they could perform. The entire system operated autonomously and without intervention until the mission's completion. The external environmental factors affecting the platform's stability and mission success during execution, such as sea state, light conditions, temperature, and humidity, were unknown. Additionally, the positions and orientations of the multiple objects to be moved were randomized. These conditions closely mirror the demands of real-world marine cargo transport scenarios, where systems must be capable of quick deployment upon cargo arrival, with no opportunity for repeated testing.
{\colc While challenging,} the deployment of robotic systems in genuine maritime environments, under conditions similar to those encountered in actual cargo transport scenarios, presents an exceptional opportunity for developing a truly reliable robotic system.

Motivated by the above requirements and challenges, this paper presents a comprehensive design of an autonomous aerial system.
The system transport cargo from a target deck to an Unmanned Surface Vehicle (USV), serving as a landing platform, in a maritime environment without GNSS. The primary objectives of the system include accurately localization the UAV platform and target cargoes, establishing reliable target adsorption capacity, and ensuring successful transportation to the landing platform. 
The main contributions of this paper can be summarized as follows:

\begin{itemize}
	\item{\colo In a GNSS-denied maritime environment, a landmark-based localization scheme was developed. {\colm This scheme utilizes ultrawideband (UWB) and QR code facilities deployed on the landing platform to provide the UAV with location information tailored to varying performance needs.} Additionally, it addresses the problem of misalignment between the UAV and landing platform coordinate systems due to magnetometer interference.}
	\item{\colo A mechanical adhesion mechanism was designed specifically for cargo simulated by a suitcase. This adhesion system primarily consists of a rail, carbon board, sponge, and adhesive tape. It is driven by an onboard screw motor to ensure secure attachment to the cargo.}
	\item{\colo A vision-based servo landing method for UAV was designed. This approach utilizes a fixed onboard camera to provide the cargo's position. By employing outlier rejection and smoothing filters on the detected data, the method effectively mitigates oscillations caused by cargo movement or uncertain external disturbances.}
	\item{\colo The developed autonomous system underwent numerous iterations and parameter optimizations through extensive outdoor experiments, enhancing its feasibility and robustness for performing transport tasks in marine environments. It successfully completed the mission during the competition, earning a feature report from the event organizers\footnote{https://youtu.be/sRkTnEPF9S0?si=1nIFA$\_$9Qy$\_$0UClp1}.}
\end{itemize}

\section{Related Work}
{\colc Aerial transportation systems are gradually being applied more extensively in maritime transportation, such as search and rescue missions and goods delivery. Specifically, these systems can be used to deploy satellite phones, medical kits, and consumables \cite{ref:MAJ2011}.
Currently, humans have already participated in real maritime aerial transportation tasks using aerial platforms. For instance, emergency medical assistance has been carried out via helicopters near the coast of France \cite{ref:PHA2021}. The challenges of pickup and delivery in maritime logistics operations using helicopters have been studied in \cite{ref:RAT2007}.}

\subsection{Aerial Autonomous Transport System}
Recently, UAV offer a solution for faster and more efficient delivery of cargo, thereby cutting down on transport costs and reducing carbon emissions, as evidenced by recent research \cite{ref:AAJ2023}.
Considering autonomous systems, aerial transport platforms typically leverage a combination of onboard sensors, grasping mechanisms, and pre-defined algorithms to navigate and execute cargo delivery missions.
An off-centered aerial grasping system is developed in \cite{ref:ADI2024}, consisting of a localization system, visual detector, thrust controller, grasping mechanisms, and an end-to-end state machine.

Different tasks contribute to varying aerial transport system designs. For instance, a UAV with two independently controlled tilting arms was developed in \cite{ref:CAI2021} to enable more flexible grasping in the air. Based on this structure, a dual-level adaptive robust control system was designed to manage the uncertainty of inertial parameters.
{\colc In the maritime industrial application of cargo transportation by UAVs, \cite{ref:FSI2023} developed a sliding mode controller to handle external disturbances. Based on the Gazebo simulation system, they simulated the process of a UAV transporting cargo from one ship to another.}
From cargo intervention perspective, the design of a UAV clamping solution often depends on the target's characteristics such as shape, weight, and material. In \cite{ref:DDI2011}, the authors designed a low-complexity, lightweight electromagnetic gripper to pick up and transport wood-based cargoes.
A multi-propeller-based aerial manipulator, featuring a robotic arm with 7 degrees of freedom, was demonstrated by grasping a structural bar in \cite{ref:GCI2014}. Additionally, \cite{ref:LAI2020} developed a suction cup-based manipulator for UAVs to perch or grab targets, which requires a flat contact surface to provide adhesion.

\subsection{UAV in GNSS-Denied Environment}
In GNSS-denied environments, UAVs commonly rely on self-localization achieved through analyzing image streams captured by one or more onboard cameras, such as visual simultaneous localization and mapping (VSLAM) \cite{ref:CPI2016}.
In \cite{ref:ADI2024}, the authors introduce an indoor cargo transportation UAV utilizing a GPU-accelerated stereo VSLAM method.

Indeed, limitations persist in complex light conditions and environments lacking distinct structures, which can potentially compromise the performance of VSLAM.
In GNSS-denied navigation scenarios, lidar-based approaches \cite{ref:JLI2020} offer promising capabilities in obstacle avoidance and pose estimation enhancement.
Moreover, hybrid solutions such as the visual-lidar method \cite{ref:XLI2020} have garnered significant attention for their ability to mitigate sensor degradation issues.
Despite the growing maturity of vision- and lidar-based autonomous navigation methods, they still face challenges when confronted with weakly textured environments in the oceans and the scattering of light by the water surface.

{\colt UWB technology has emerged in recent years as an effective localization technique in location-aware sensor networks, which offers highly accurate ranging capability \cite{ref:SMI2010}.
In addition, one of the advantages of bidirectional UWB ranging and communication is its ability to simultaneously serve both navigation and tracking tasks as needed in specific applications \cite{ref:SBI2023}.}
Reference \cite{ref:DKI2020} develops an integrated indoor localization system combining IMU and UWB, utilizing EKF and UKF to improve system robustness and accuracy.
{\colt The UWB modules on the market supports real-time positioning with high refresh rates, making it ideal for dynamic environments where UAVs are in constant motion and have small size and weight \footnote{https://www.nooploop.com}, making it easy for UAVs to mount.}
A relative localization and navigation method for docking UAV and mobile platforms is developed in \cite{ref:CAJ2022}, utilizing UWB and vision integration to enhance relative position estimation accuracy.
{\colt Due to its wide frequency range and low power spectral density, UWB is less susceptible to interference from other wireless technologies, ensuring stable and reliable communication \cite{ref:DTI2002,ref:NUI2020}.
More importantly, compared to lidar-based \cite{ref:JLI2020} or vision-based \cite{ref:XLI2020} SLAM localization techniques, UWB, as a radio technology, is less affected by surface reflections or dynamic textures, making it more suitable for UAV positioning in maritime environments.}
{\colo In the localization algorithm, the UWB facilities primarily provide distance measurements between nodes. Beyond the previously mentioned work, there is extensive research on distance-based cooperative relative localization and navigation for multi-UAV systems \cite{ref:RCI2022,ref:BRA2021,ref:ASJ2019}.}
{\colm In the aforementioned scenarios, the UWB module is applied directly, assuming stationary UWB anchors. However, none of these approaches consider non-stationary UWB anchors.
This paper emphasizes the challenges of employing UWB-based localization method in marine environments.}

\subsection{Challenges Faced}
Upon reviewing related work, it becomes evident that existing transportation solutions have successfully utilized UAVs to independently track designated targets and execute precise cargo delivery operations. However, much of this work is confined to controlled laboratory conditions with stable environments, thus lacking verification in real-world dynamic outdoor environments, particularly in marine environments where GNSS is denied. In such environments, factors such as strong winds, turbulence, and other disturbances can significantly impact the transportation efficiency and stability of aerial transport systems. Therefore, it is crucial to comprehensively analyze the components of the handling system as they are affected by real-world conditions and to expand them based on the task requirements. Additionally, the interdisciplinary nature of system development and integration poses a significant challenge.

{\colo To address the challenges of the specific environment and the tasks in MBZIRC 2024, we proposed several innovative solutions based on above existing work.
In terms of transportation, we designed a novel attachment mechanism equipped with a camera for target recognition. The onboard motor indirectly drives the camera's vertical movement, allowing the UAV to have a wider field of view from a certain height. This enables effective cargo identification at lower height, improving the UAV's landing accuracy and the reliability of cargo attachment.
For localization, we deployed UWB facilities on a non-static platform (landing platform) and equipped the UAV with dual UWB labels to establish an attitude reference relative to the landing platform. This approach decouples the impact of the landing platform's attitude oscillations at sea from the UAV's flight control, preventing the UAV from oscillating in sync with the platform's movements.}

\section{Transport System Overview}
The design of the system is motivated by the mission requirements of MBZIRC2024, with the primary objective being to transition the autonomous capabilities of the UAV from the laboratory to a real-world environment. The following provides a concise overview of the aerial transport system in terms of task flow and modular composition.

\subsection{State-Driven System for Cargo Transportation}
\begin{figure}[t]
    \centering
    \includegraphics[scale=1]{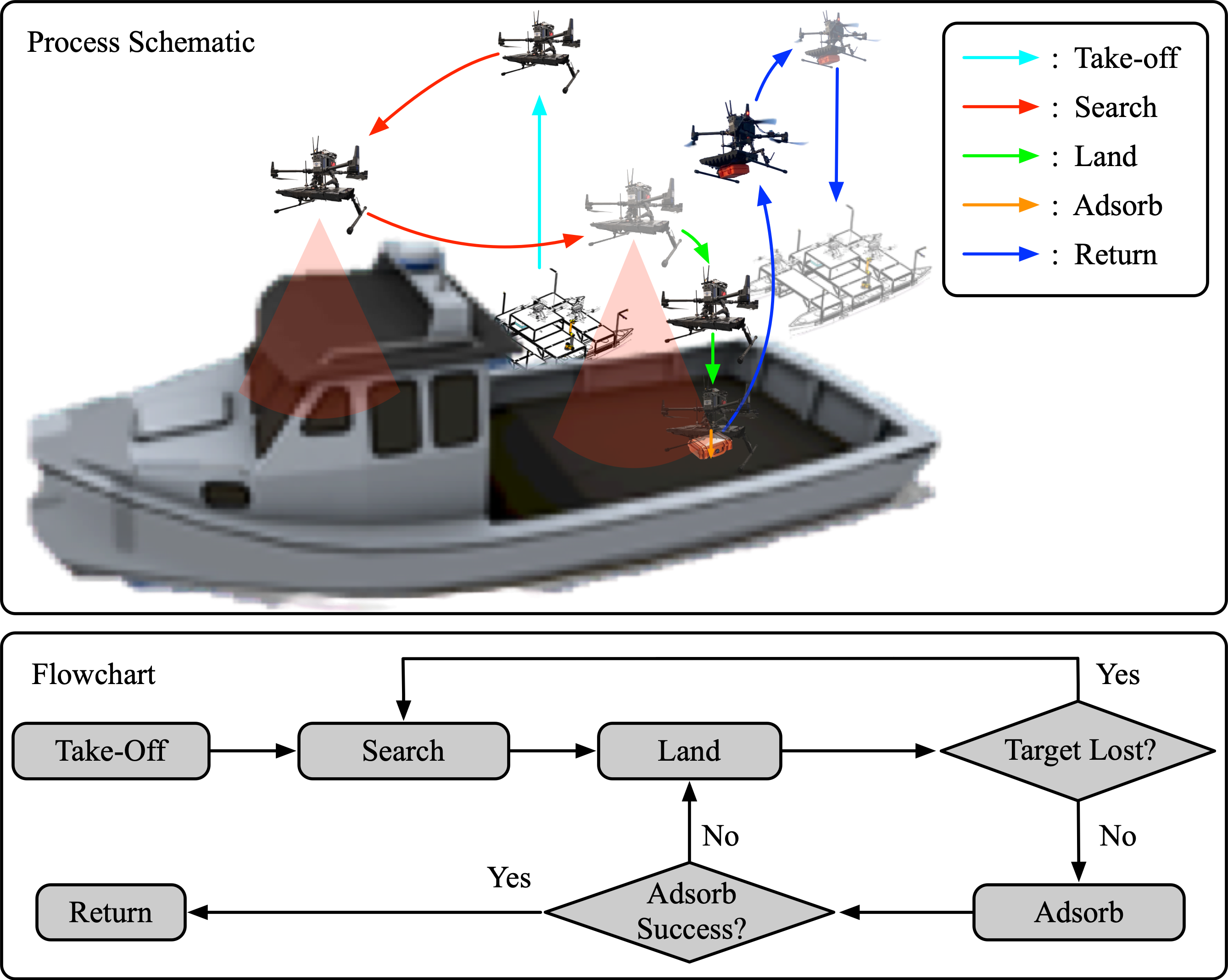}
    \caption{The schematic diagram of the state machine for the autonomous aerial transport process.}
    \label{fig:state_machine}
\end{figure}

To ensure the entire task can be autonomously operated, the UAV system utilizes a state machine to coordinate the cargo transportation process. This state machine ensures a logical sequence of operations, guiding the UAV through various flight phases in an autonomous manner. These phases include take-off, search, visual servoing and landing, manipulation, and return, as illustrated in Figure \ref{fig:state_machine}.

\begin{itemize}
    \item \textbf{Take-off}: The system transitions through distinct flight phases based on critical events and sensor data. The first phase involves take-off. Upon reaching the designated deck area, the system receives a signal and initiates the take-off phase. During this phase, the QR visual localization module guides the UAV's ascent. Given that the landing platform may be in motion, the system prioritizes horizontal tracking to prevent collisions with the platform or its equipment.

    \item \textbf{Search}: Once the UAV reaches the predetermined height, it enters the search phase. During this phase, the UAV traverses a pre-defined path using UWB localization for moving. If the target cargo remains undetected after covering the entire deck area at the current height, the system initiates a descent maneuver. This descent aims to improve image resolution and enhance target recognition by the CNN network within the perception module.

    \item \textbf{Land}: Target acquisition triggers the visual servoing phase. Here, the perception module, utilizing the camera data, tracks and guides the UAV towards the cargo for a precise landing. Upon reaching a critical height where the cargo fills the camera's field of view (hover directly above the target), a rapid landing maneuver is executed.
    
    \item \textbf{Adsorb}: After a successful landing, the manipulation phase commences. A carbon board with waterproof adhesive descends and adsorbs to the target cargo via controlled motor rotation, facilitated by the manipulation module. Following a brief stabilization period to ensure a secure attachment, then the UAV takes off.

    \item {\colr \textbf{Return}: After takeoff, the UAV hovers to determine whether the cargo has been successfully attached based on the current thrust. If the attachment fails, the UAV returns to the landing phase to attempt reattachment. If the attachment is successful, the  UAV returns to the takeoff platform, relying on UWB localization for navigation. As the UAV approaches the landing platform and the QR code comes into the camera's field of view, it switches to QR-based localization for landing. At this point, the entire transportation process is completed.}
\end{itemize}

It is worth noting that the system can incorporate brief hovering periods before and after cargo manipulation. By recording the average rotational velocity of the rotors during these hovering phases, the system can calculate the thrust difference pre- and post-adhesion. This data, analyzed by the onboard computer, serves as an indicator of successful cargo attachment. Based on this information, the system can make informed decisions regarding subsequent actions, such as attempting to re-capture the cargo if necessary.

\subsection{System Components}
{\colt In marine GNSS-denied environments, transporting cargo from the deck of a target ship to a landing platform requires not only the UAV platform, but also various functional modules to support the autonomous operation of the entire system.
{\colc In GNSS-denied maritime conditions, the transportation system adopts a landmark-based localization solution. QR codes, as a typical artificial landmark, are widely used for UAV localization due to their low cost and ease of implementation \cite{ref:TAD2024,ref:KEI2022}.
Compared to QR code landmarks, UWB, a radio-based technology, provides a larger localization range and is also widely used in UAV navigation \cite{ref:AAJ2013,ref:CAJ2022}. The localization module integrates both methods to reliably and stably support the UAV's navigation tasks.}
{For small object detection, CNN-based methods currently offer efficient and fast solutions \cite{ref:KRI2020}. The perception module integrates a pre-trained cargo detection model to provide the pixel location of the cargo in the image, assisting the drone in landing.}
The path planning module is responsible for executing the UAV's search tasks within the landing platform's coordinate system, ensuring coverage of the target ship's deck area based on the detection range.
The autonomous flight control module calculates real-time control commands based on the current state, driving the UAV to converge to the designated position {\colc based on DJI Open Source Development Kit (OSDK)\footnote{https://developer.dji.com/onboard-sdk}}.
The operation module is primarily responsible for controlling the motors to drive the adhesion mechanism, enabling the attachment of the cargo.
{\colc In fact, references \cite{ref:WFC2021,ref:HUI2020} also investigated adhesive-based gripping methods, developing an anthropomorphic end effector and an underactuated adhesive gripper, respectively.}
In addition, communication between the functional modules is facilitated by the {\colc Robot Operating System (ROS) \footnote{https://www.ros.org}}, as shown in in Figure \ref{fig:component}.
{\colc The implementation framework for target tracking control on the DJI M300 platform using OSDK and ROS can refer to \cite{ref:AHI2021}.}}
Next, we present the specific components of each module and their corresponding design concepts:

\begin{figure}[t]
    \centering
    \includegraphics[scale=0.5]{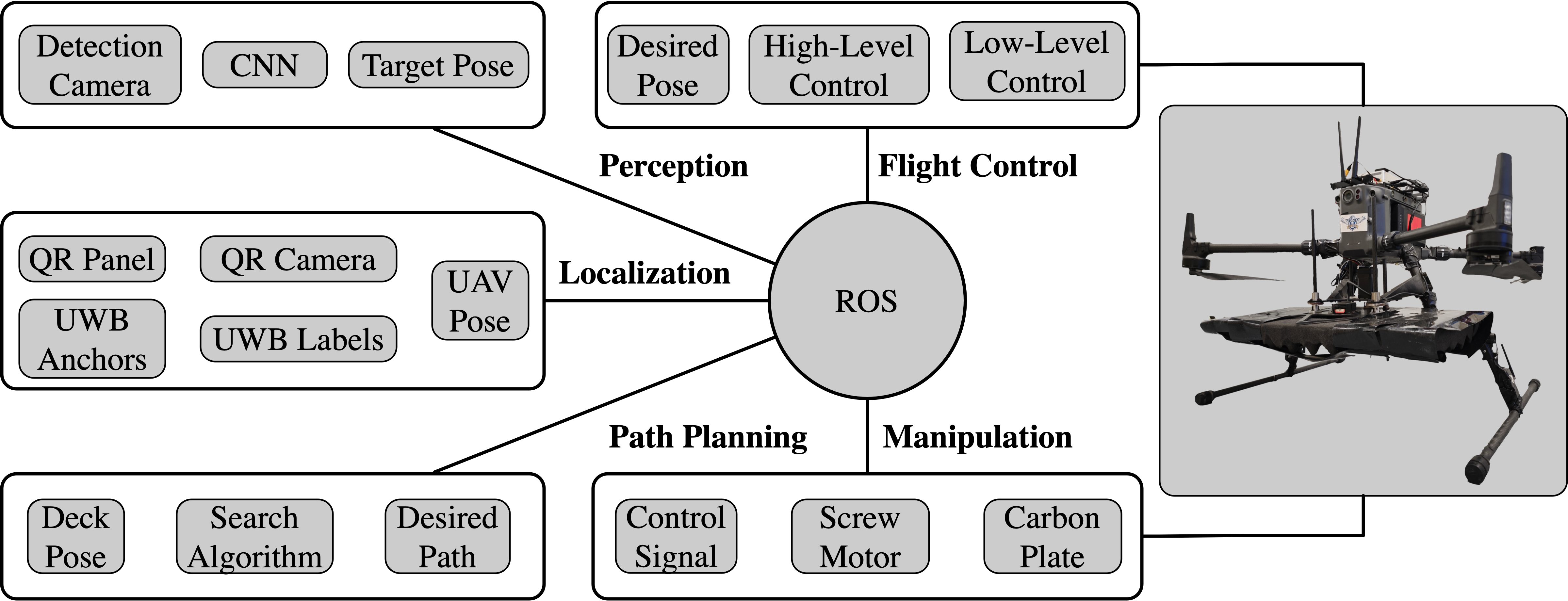}
    \caption{The system components of autonomous aerial transport system.}
    \label{fig:component}
\end{figure}

\begin{itemize}
    \item \textbf{Localization}: 
    {\colt Due to the influence of sea surface textures and the moving platform, vision-based odometry methods \cite{ref:CVJ2009} struggle to provide effective localization for UAVs.
    To provide reliable pose estimation for the UAV in the specific environments, the localization module integrates two solutions: QR-based and UWB-based localization methods.
	In the QR-based method, the landing platform is equipped with a matte-finished hardboard printed with multiple QR codes. The UAV, using a downward-facing camera, recognizes the QR codes and extracts the corresponding position encoding information, allowing it to determine its pose relative to the landing platform.
	When the QR codes are out of the camera's field of view, the QR-based localization method becomes ineffective, and the module switches to UWB-based localization. This method calculates the drone's position using distance measurements from multiple nodes. To enhance the observability of the position, six UWB anchors are installed on the landing platform. Additionally, two UWB tags are mounted on the drone to obtain its relative heading with respect to the landing platform.
Additionally, since the landing platform is subject to swaying on the sea surface, the module needs to subscribe to the platform's attitude angles to decouple pose errors caused by the motion.
    }

    \item \textbf{Perception}: {\colt The cargo to be transported is placed on the deck of the target ship with unknown location. The perception module is responsible for identifying the designated cargo and providing its relative position to guide the UAV's landing.
   	By balancing the overall payload weight of the UAV, the perception module utilizes a fixed wide-angle industrial camera to capture images of the cargo, rather than opting for a more flexible gimbal camera solution \cite{ref:LRA2020}.
	The cargo are designated as small suitcases of different colors. Initially, they were placed in various backgrounds and lighting conditions for image collection, which was then used to train a convolutional neural network (CNN) model, resulting in a pre-trained cargo detection model.
	To reduce the impact of background clutter on detection accuracy, the region of interest (RIO) strategy was further employed \cite{ref:WSI2019}. This involves cropping a fixed proportion of the video frame around the cargo, which is then used as input for the detection model.
	Once the cargo is reliably detected—identified as the same object over multiple consecutive frames—its pixel location in the image is transmitted to the flight control module to provide feedback for landing control.
    }

    \item \textbf{Path Planning}: 
    {\colt The UAV's detection range is constrained by the camera's field of view and flight altitude, preventing full coverage of the area where the cargo may be located, that is the target ship's deck.
    The path planning module aims to provide a set of waypoints that guide the UAV to cover the entire deck of the target ship. To efficiently cover the area, the planning module must first know the center position and heading of the target ship in the landing platform's coordinate system, as well as the relative orientation  and size of the deck.
    The above information is assumed to be known in this paper. In practical scenarios, the position and orientation can be obtained through point clouds dataset collection and computation by mounting a LiDAR on the landing platform \cite{ref:ZAI2024}.
}
 
    \item \textbf{Autonomous Flight Control}: 
{\colt The flight control module primarily calculates control commands for different stages of the transportation process, driving the UAV to converge to the desired pose. Specifically, the module implements a two-tier control system. The upper-level controller, embedded in the onboard computer, calculates the desired velocity commands in the UAV's body frame, storages and switches specific control parameters for each process. The lower-level controller, embedded in the DJI platform, receives velocity commands from the upper-level controller and drives the motors to follow these commands.}
In addition, communication between these two level controllers is facilitated through the DJI OSDK.

    \item \textbf{Manipulation}: 
{\colt The manipulation module is responsible for connecting the cargo to the UAV and transporting it to the landing platform. Considering the deck movement of the target ship and wind interference, the manipulation task is executed after the UAV has stabilized its landing, which increases the probability of successful cargo capture. The medium for connecting the UAV and the cargo is a waterproof adhesive tape, which is attached to a carbon plate connected to the UAV. The carbon plate is driven by a motor and a specially designed guide rail to adhere the cargo. The motor's movement commands are issued by the onboard computer, primarily based on whether the UAV has stabilized its landing.
}
    
\end{itemize}

\section{Modular Hardware Platform}
This section delves into the hardware-level design of the aerial transport platform, breaking down the platform into modular components. It provides insights into the characteristics of the corresponding hardware and outlines the design ideas, aiming to accelerate research in related fields.

\subsection{Platform Description}
\begin{figure}[thpb]
    \centering
    \includegraphics[scale=0.05]{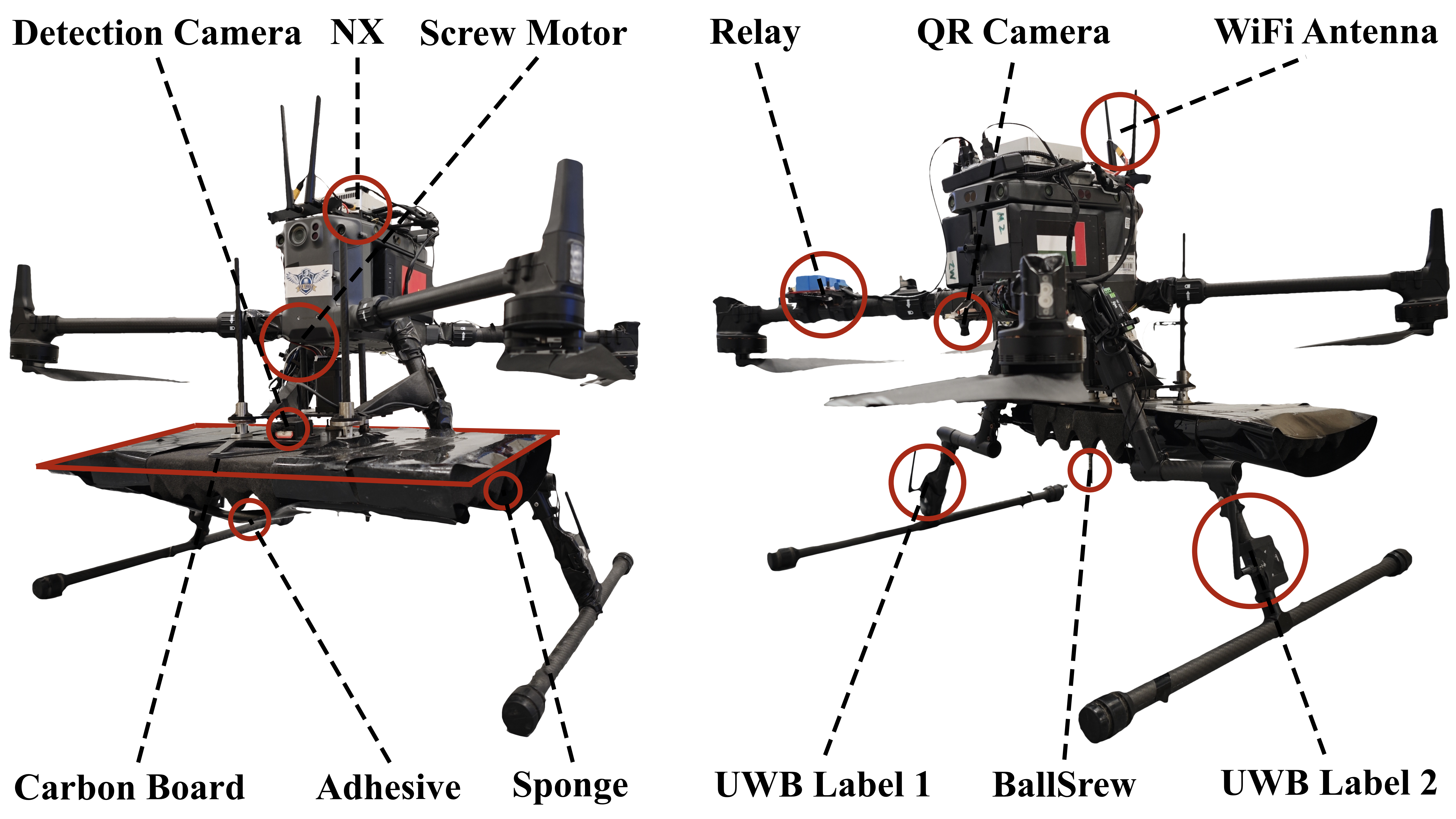}
    \caption{The proposed GNSS-denied aerial transport platform.}
    \label{fig:platform}
\end{figure}

In order to ensure the safe execution of outdoor experiments and expedite the development of aerial cargo platforms, we conducted research using the commercial quadrotor platform DJI M300, {\colc which also are utilized to build a forest aerial fire detection system in \cite{ref:LAI2022}.} The platform offers an additional $2.7$kg of load capacity {\colc and the specific technical specifications can be found in link\footnote{https://www.dji.com/support/product/matrice-300}.} The onboard GNSS modules were removed from the platform to comply with race regulations and operate in a GNSS-denied environment.
Furthermore, we designed structural components using 3D Computer-aided Design (CAD) models to secure the onboard devices for localization, perception, and computation of the transport system. An installation schematic is depicted in Figure \ref{fig:platform}, and quantitative metrics related to the tasks are presented in Table \ref{tab:platform}.
{\colb With the integration of all onboard modules—including the computing unit, sensors, and adhesion structure—the total weight of the aerial transport platform reaches approximately 7.9 kg. This configuration allows for an additional payload capacity of around 1.2 kg for the cargo.}
In addition, the DJI flight controller provides the OSDK to users, offering access to the IMU data of the quadrotor and enabling reception of velocity commands. This research primarily focuses on localization, detection, and high-level control algorithms, which are deployed within the onboard computer. It does not delve into the low-level flight control of the aerial transport platform.

\begin{table}[!htbp] 
\centering 
\vspace{5pt} 
\caption{The device specifications of platform} 
\begin{tabular}{c|c|c} 
\hline 
Device & Attribute & Specification \\
\hline
\multirow{4}{*}{\makecell[c]{Drone\\{\colc (DJI M300)}}} &Size &810$\times$670$\times$630 mm\\
&Weight &6.3 kg\\
&Dead Weight &9 kg\\
&Hover Time  & 55 min\\
\hline 
\multirow{4}{*}{\makecell[c]{UWB\\{\colc (LinkTrack P-A)}}} &Range & 40 m \\
&Frequency &4/4.5/6.5GHz\\
&Accuracy &10 cm\\
&{\colb Height} & {\colb 43.2 g}\\
\hline
\multirow{4}{*}{\makecell[c]{QR Camera\\{\colc (RER-USB12MP01)}}} &Resolution  &1920$\times$1080 pixel\\
&Frequency  &120 Fps\\
&Focal Length & 3.6 mm\\
&{\colb Height} & {\colb 42.6 g}\\
\hline
\multirow{4}{*}{\makecell[c]{Detection Camera\\{\colc (RER-H200S)}}} &Resolution  &1280$\times$720 pixel\\
&Frequency  &60 Fps\\
&Focal Length & 2.7 mm\\
&{\colb Height} & {\colb 60 g}\\
\hline
\multirow{4}{*}{\makecell[c]{WiFi\\{\colc (Intel AX210)}}} &Frequency  &2.4/5/6 GHz\\
&Tx Power &18 dBm\\
&Antenna Gain &6 dBi\\
&{\colb Height} & {\colb 8.4 g}\\
\hline
\multirow{4}{*}{\makecell[c]{Onboard Computer\\{\colc (NVIDIA Jetson Orin NX)}}} & CPU & 8 Cores, 2 GHz\\
& GPU & 32 Cores, 918 MHz\\
& Memory & 16GB LPDDR5\\
&{\colb Height} & {\colb 188 g}\\
\hline
\multirow{3}{*}{\makecell[c]{Screw Motor\\{\colc (FOXON Customized)}}} & Holding Torque &0.7 Nm\\
& BallSrew Length & 155 mm\\
&{\colb Height} & {\colb 600 g}\\
\hline
{\colb Other Payloads} & \multirow{3}{*}{\colb Height} & \multirow{3}{*}{\colb 459 g}\\
{\colb \makecell[c]{(carbon board, brackets,\\ wire materials, etc.)}} & &
\end{tabular}
\label{tab:platform} 
\end{table}

\subsection{Localization Facilities}
\begin{figure}[thpb]
    \centering
    \includegraphics[scale=0.2]{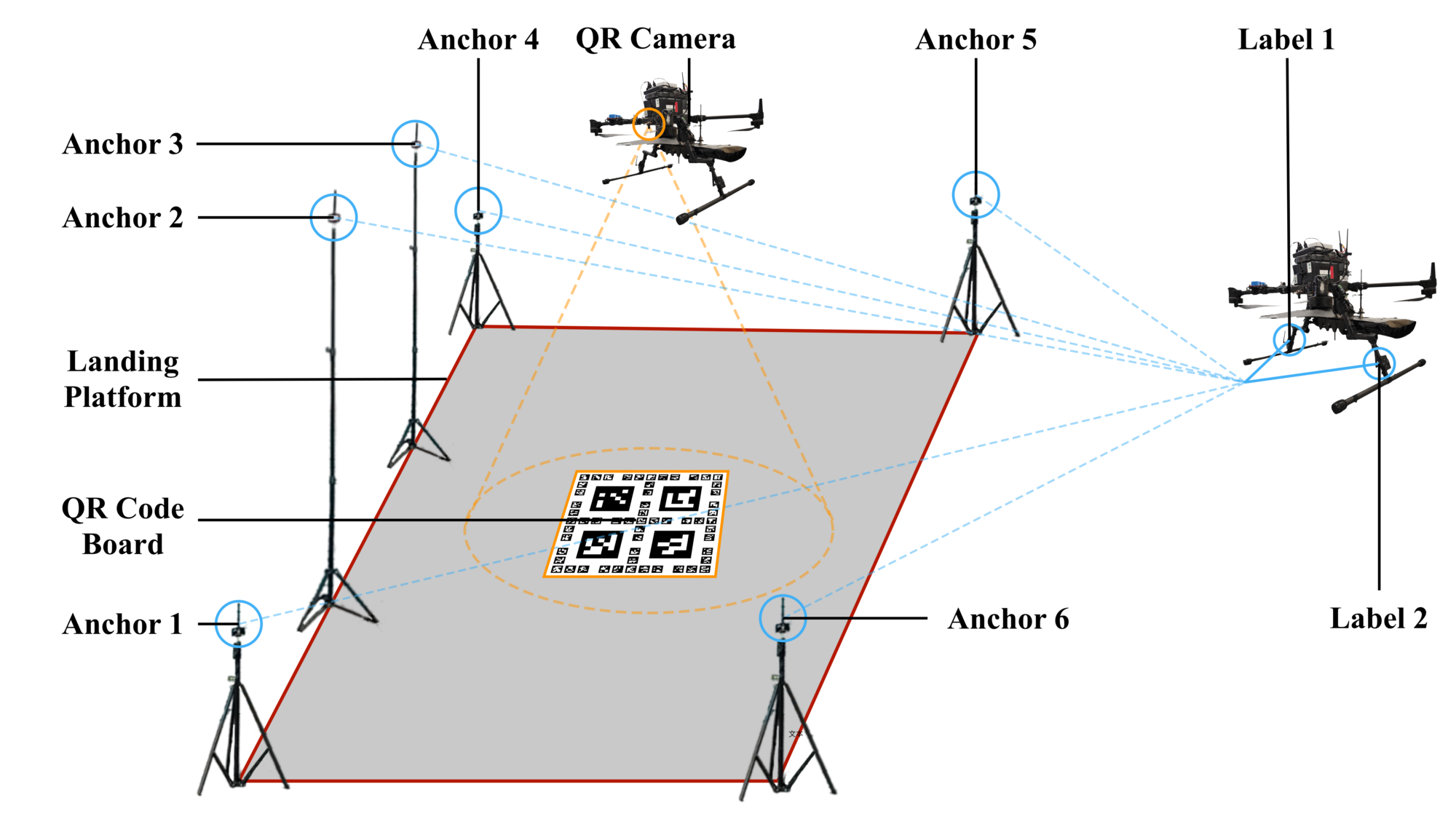}
    \caption{\colr The schematic diagram of the localization scheme.}
    \label{fig:localization}
\end{figure}

The dynamic and sparse characteristics of the ocean surface present significant challenges for deploying laser radar on UAVs \cite{ref:BOO2019}.
{\colc Due to the constraints of the overall mission environment, this paper adopts a landmark-based localization method, which includes both QR and UWB anchors, fixed on the landing platform. Assuming the platform is a flat rigid board equipped with an IMU to obtain its three-axis attitude angles in inertial space, we set the center of the platform as the origin of the inertial coordinate system. Using the fixed relative positions of the QR and UWB anchors on the platform, along with the attitude angles of the platform, we can calculate the coordinates of the anchors in the inertial coordinate system. Furthermore, once the position of the UAV relative to the anchors is obtained, we can determine its position in the inertial coordinate system, allowing the UAV to execute navigation tasks within this reference frame.}
An illustration of the anchors's installation and localization scheme can be found in Figure \ref{fig:localization}.

{\colc The QR code localization method, due to its high precision after calibration, is primarily involved in the UAV's takeoff and landing tasks.
In this method, QR codes are affixed to designated areas for landing platform, which can be generated on demand using a website called the online barcode generator\footnote{https://barcode.tec-it.com}.}
Acknowledging the challenge of camera overexposure in outdoor settings, it is imperative to print QR codes on matte surfaces. Furthermore, to accommodate precise localization of the UAV at different heights, multiple QR codes of varying sizes are arranged on a single board.
{\colc The information fusion problem in the detection of multi-sized QR codes has been studied in \cite{ref:ZDI2022} and applied in ground mobile robots\cite{ref:ZLI2022}.}
An industrial camera, called QR camera , is mounted on the tail of the UAV to recognize the QR codes on the takeoff and landing platform.
For precise localization and to minimize image preprocessing, it's necessary to utilize an undistorted, gray-scale camera.
It's important to emphasize that manual focusing of the camera is essential to ensure clear QR code images are obtained at varying heights.
By conducting calibration, the camera can offer an approximate horizontal field of view of 81 degrees and a vertical field of view of 53 degrees.

{\colt Due to the limited field of view of the QR camera, when the QR code is out of sight, the UAV will utilize the UWB-based localization for navigation, particularly during the search and return phases.
The placement of UWB anchors determines the UAV's locatability within its movement space. Since UWB localization essentially estimates the UAV's position states using ranging information, according to reference \cite{ref:TPA2017}, at least three non-collinear fixed anchors are required to ensure the observability of three-dimensional position states.
Moreover, the spatial configuration of UWB anchors can significantly affect the system's localization accuracy. The optimal configuration methods of anchors's locations can be found in \cite{ref:FAI2021,ref:CAK2024}.}

{\colr According to above references, localization accuracy depends not only on the spatial configuration of the anchors but also on the position of the UWB labels. Compared to the internal space enclosed by multiple anchors, UWB labels have lower localization accuracy in the external space. Considering that the UAV will operate outside the landing platform, we aim to maximize the enclosed space. To enhance the reliability of the localization system, a total of six UWB anchors were selected. Four of them are positioned at the four corners of the landing platform to expand the enclosed space, and their height is set low enough to reduce the risk of collisions. The remaining two anchors are placed on the same side, and the UAVs will across another side to minimize the chance of collisions.
In addition, these two anchors are positioned at the higher fixed points, improving the vertical localization accuracy.
}
To mitigate potential interference with UWB signals, the UWB labels can be affixed to the UAV's landing gear. This placement is advantageous as, during mission execution, the UAV typically remains positioned above the anchors in most scenarios.
Prior to take-off, as the UAV is situated on a moving platform at sea, its attitude undergoes constant fluctuations. This results in significant yaw drift, hindering the establishment of a unified coordinate system with the landing platform. Consequently, the UAV is equipped with two fixed-position UWB labels to calculate the relative yaw angle between itself and the landing platform.

\subsection{Detection Camera}
In this mission, the location of the target cargo is unknown, and the accuracy of UWB localization is contingent on distance, thus indirectly influencing the UAV's search range. To enhance search efficiency and mitigate risks associated with UAV operations over the sea, rapid target acquisition coupled with vision-based UAV control proves highly advantageous for transport tasks.

{\colc To visually capture the cargo efficiently, the ideal scenario is for the UAV to navigate directly to the cargo area after takeoff. One approach is to use a gimbal-mounted camera to search for the target from above the launch platform \cite{ref:LRA2020}, but due to the UAV's weight limitations, this solution is impractical. Alternatively, a wide-angle camera with a larger field of view can be employed to search over the deck of the target vessel until the cargo is found. If the relative position between the UAV and the UAV's deck is known, the search area can be significantly reduced. To enable this, we deploy a LiDAR\footnote{https://www.livoxtech.com/mid-360} on the launch platform to estimate the pose of the vessel's deck and provide this information to the UAV, thereby minimizing the search area and improving mission efficiency.
It is worth mentioning that, compared to a gimbal camera, using a fixed-mounted camera to acquire the cargo's position introduces attitude coupling between the camera and the UAV. This coupling presents certain challenges for the landing control of the UAV, as any changes in the UAV's attitude can affect the accuracy of the cargo's position estimation.}

It's worth noting that a wider field of view for the detection camera extends the search coverage range at a certain flight height. However, a broad field of view often introduces distortion in camera imagery, complicating cargo identification.
{\colc To determine the boundaries of image distortion for this task, we first evaluated the recognition distance for a given cargo using completely undistorted images. We established that stable recognition performance could be achieved within a 6-meter range. At this distance, we tested lenses with different focal lengths to ensure that the cargo could be consistently recognized across the entire image without affecting its generalization capability. Ultimately, we selected a lens with a focal length of 2.7mm, which offers a relatively wide field of view.}

{\colc After calibration, the camera's horizontal field of view was measured at 106 degrees, and the vertical field of view at 73 degrees. Considering that the image becomes significantly blurred during the focusing process, making it difficult to recognize the cargo, this solution uses a fixed-focus camera. The focus is preset to ensure stable recognition over the entire landing height. Outdoor lighting conditions greatly affect the cargo detection performance, so the selected camera must have an automatic exposure adjustment feature. The training dataset was further collected under varying weather, time, and location conditions, considering that the final testing environment will have unknown factors related to these conditions.}

\subsection{Manipulation Structure Design}
Typically, there are two primary cargo transport strategies for UAVs: one involves securing the cargo onto the UAV's rigid body, while the other entails suspending the payload using cables \cite{ref:DAJ2020}.
In maritime settings, the decks where cargoes are positioned often lack stability. Combined with the influence of wind, accurately aligning the cable end or gripper with the cargo becomes exceedingly challenging while the UAV hovers in the air.
Hence, our approach leans towards affixing the cargo to the UAV's rigid body once the landing process is complete, followed by transportation.

\begin{figure}[thpb]
    \centering
    \includegraphics[scale=0.5]{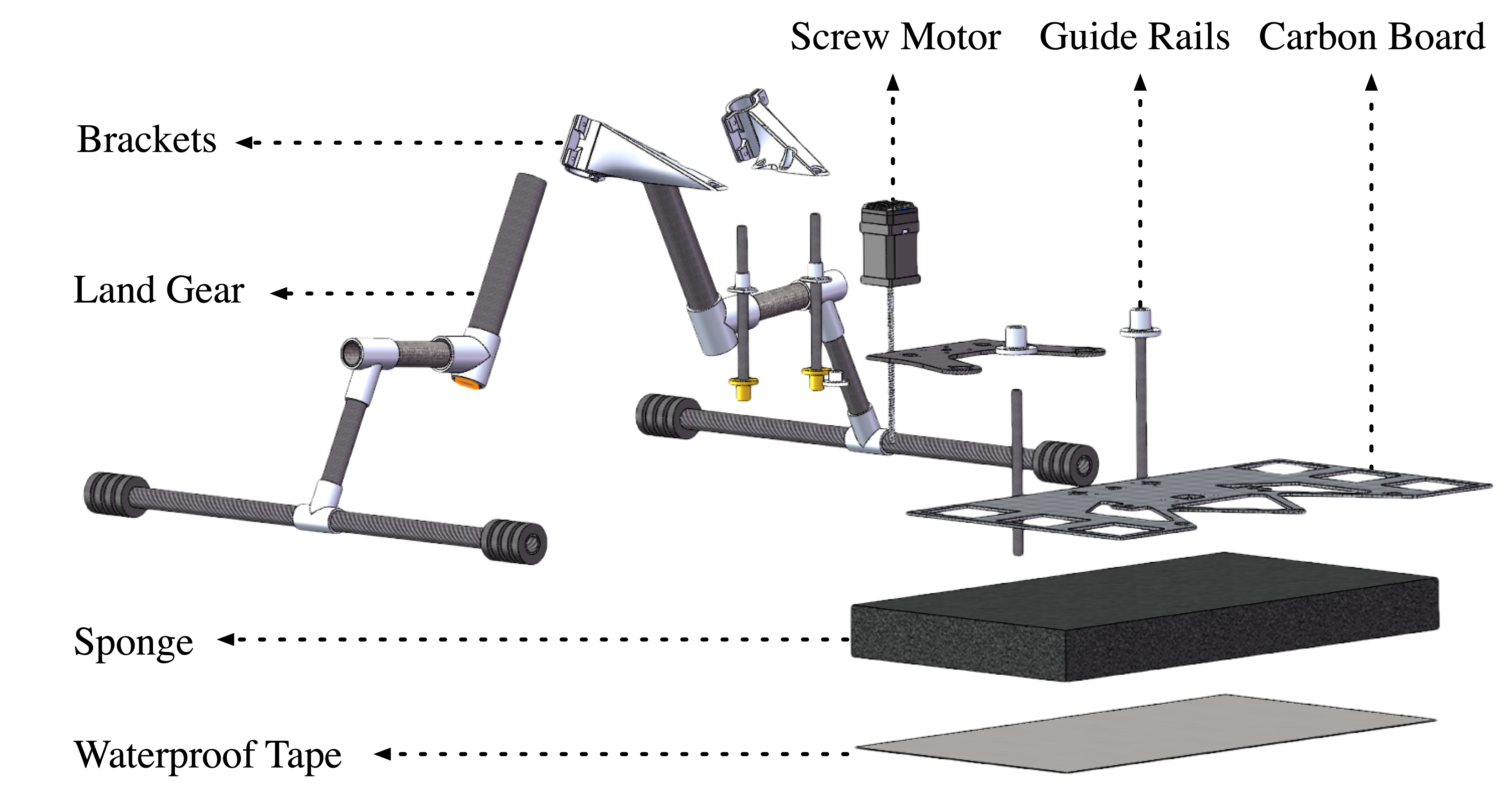}
    \caption{\colr The CAD model for manipulation structure component.}
    \label{fig:structure}
\end{figure}

In MBZIRC2024, the cargoes were enclosed in suitcases crafted from polypropylene composite resin.
Given the characteristics of the suitcase, we opted for a bonding method to adsorb the target cargo, necessitating waterproofing without leaving adhesive residue.
Consequently, we chose a waterproof tape\footnote{https://flexsealproducts.com/products/flex-tape} composed of adhesive and flexible rubber backing, capable of conforming to various shapes and surfaces, whether wet, dry, or even submerged.
Furthermore, we utilized a flat carbon board as a support for tape application. Employing a sponge layer as an intermediary buffer facilitated optimal adherence of the flexible rubber tape to the box's surface, while also shielding it from potential impact by the carbon board.
To ensure precise alignment of the tape's adhesive area directly above the cargo, we positioned the detection camera at the center of the carbon board, mitigating occlusion caused by the UAV's structure during end-stage landing, akin to the "Eye-in-Hand" approach \cite{ref:AEI2018}.
{\colc The overall structure component of the adhesive mechanism is depicted in the CAD model shown in Figure \ref{fig:structure}.}

\begin{figure}[thpb]
    \centering
    \includegraphics[scale=0.3]{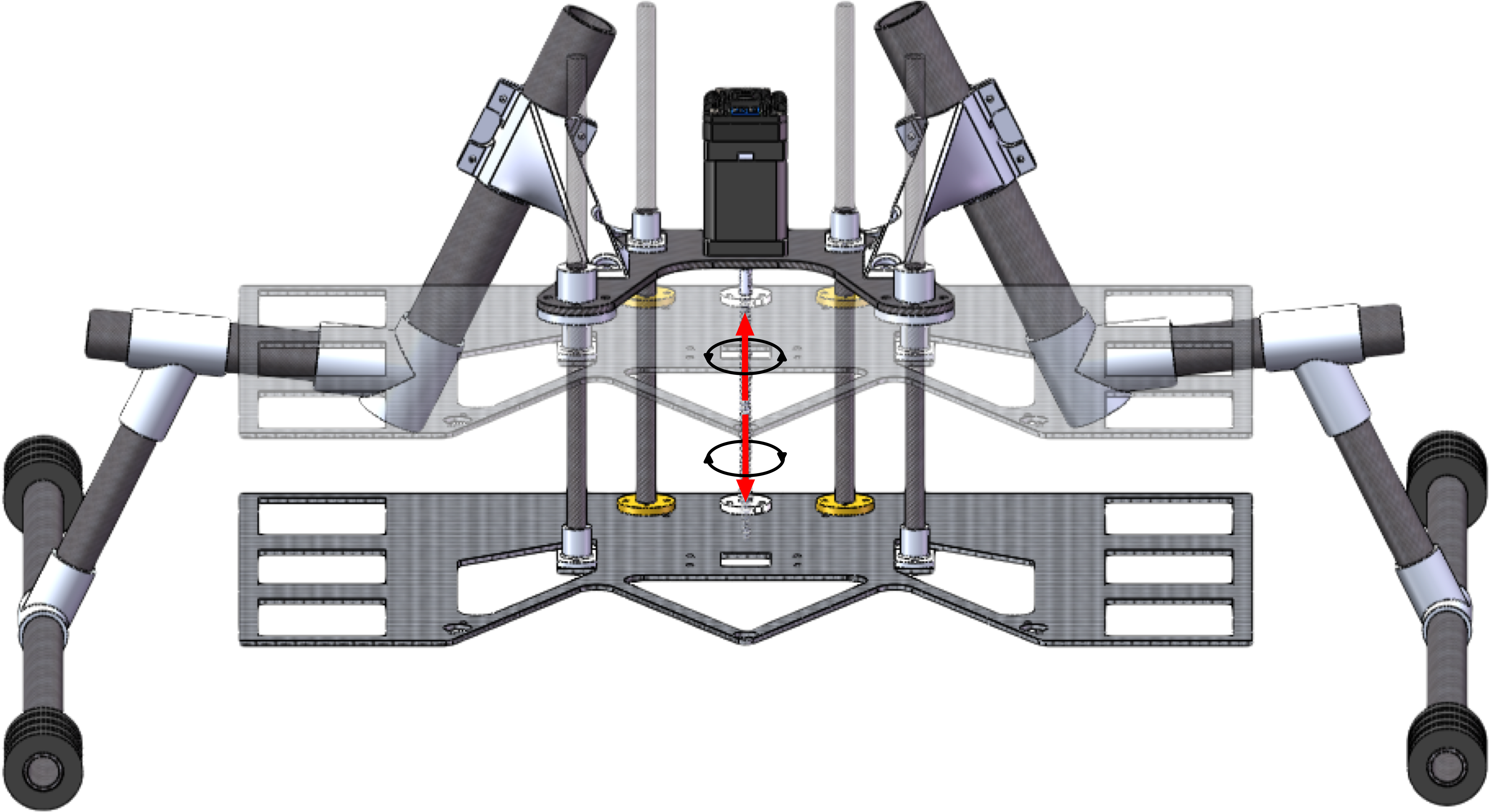}
    \caption{The illustration of carbon board's vertical movement.}
    \label{fig:gripper}
\end{figure}

Note that as the camera approaches the suitcase, limitations in its field of view may prevent the full capture of the suitcase's shape features, resulting in unstable recognition rates. Given that the UAV relies on visual recognition for control feedback during landing, unstable recognition triggers a transition to open-loop control, initiating a blind landing. Typically, this transition occurs when the suitcase fills the camera's field of view.
Moreover, in maritime environments, disturbances from deck sway and wind, coupled with ground effect and vortex ring state effects during landing, add complexity (Reference: AGI 2021). As the UAV enters open-loop control, landing errors escalate with increasing descent height.
One approach to determining the critical height for open-loop control involves elevating the detection camera during descent, achieved by raising the carbon board to which it is fixed.
To facilitate movement of the carbon board, a nut compatible with the motor lead screw is installed behind the detection camera. Four linear bearings are then affixed around the nut, along with four carbon tubes to form four guide rails. This assembly connects to the UAV body via connectors, ensuring vertical movement exclusively.
When the motor rotates the lead screw, the nut moves along its axis, thereby controlling the carbon board's vertical motion.
{\colc The schematic diagram illustrating the movement of the carbon board driven by the motor is shown in Figure \ref{fig:gripper}.}

\subsection{Communication and Kill Switch}
In the communication segment, the airborne end utilizes dual-antenna Intel AX210 Wi-Fi network cards, establishing connections with the TP-Link AX8400 router positioned on the landing platform. In this autonomous transport system, where human intervention is not feasible, the wireless communication module primarily facilitates the transmission and reception of status signals. Its main functions include receiving take-off commands and reporting the status of transport tasks.
{\colr It is noted that the onboard computer is equipped with a Wi-Fi card to enable communication between the aerial platform and external systems. This paper mainly addresses communication with the landing platform. The operation of the entire aerial transport platform is relatively self-contained, and another network protocol was employed for communication with the ground station onshore, such as Data Link, which is primarily used for monitoring purposes. Therefore, this part has been omitted from the discussion.}

Additionally, we have engineered a kill switch based on relays. This switch governs the activation and deactivation of a specific UAV motor in response to Boolean signals received from {\colc a safety observer}.
Given the GNSS-denied maritime environment, exceeding the virtual geofence prompts activation of this switch to avert potential accidents.

\subsection{Onboard Computer}
In the aerial transport system, the edge computer is powered by the NVIDIA Jetson Orin NX, offering up to 100 TOPS of computational capability to deploy autonomous algorithms for localization, detection, high-level control, state machines, and more.
The computer boasts a multitude of external interfaces facilitating seamless connection with other onboard devices, including USB 3.2, UART, GPIO, and others.

\begin{figure}[thpb]
    \centering
    \includegraphics[scale=0.5]{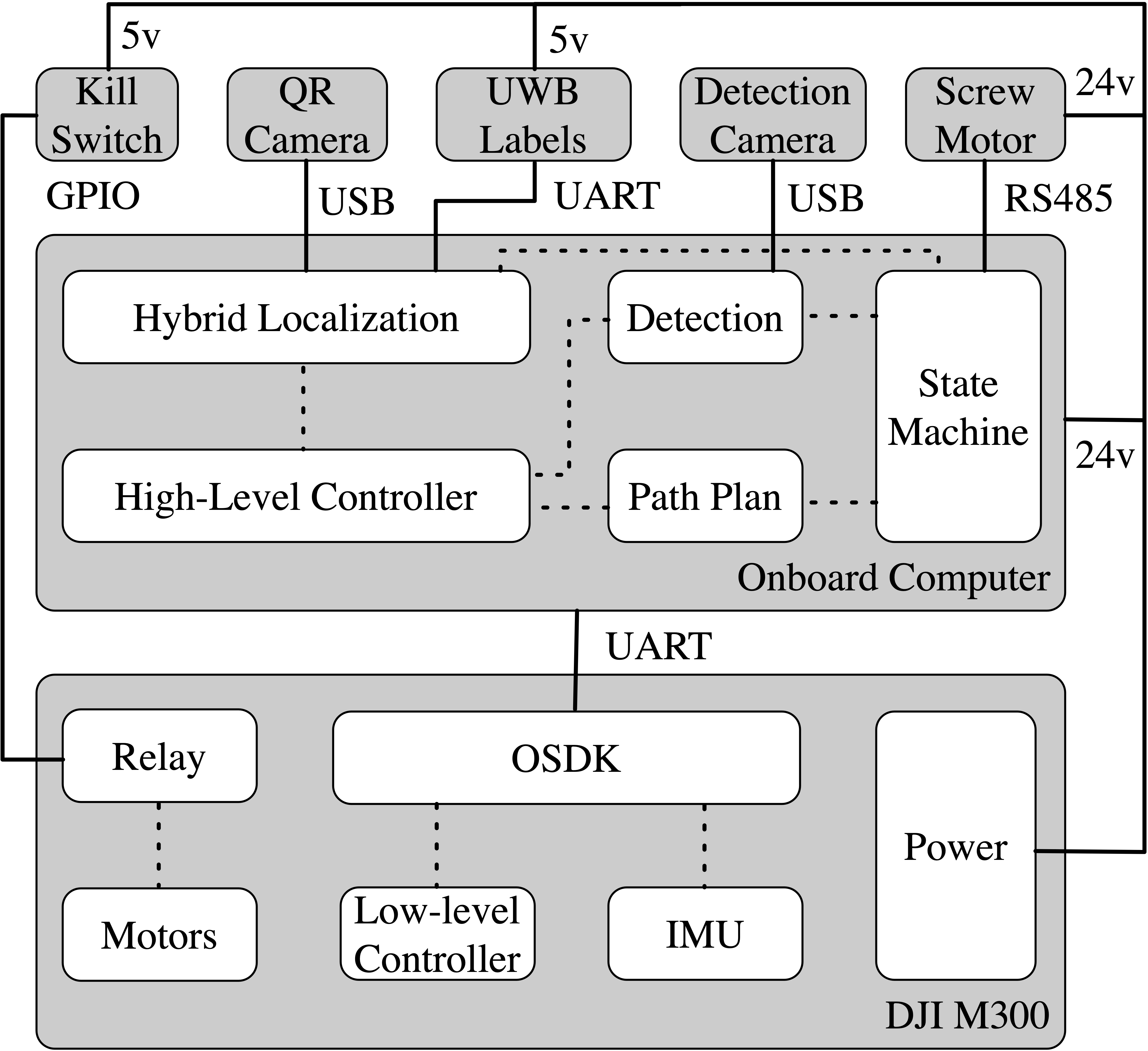}
    \caption{The block diagram for integrated connectivity of aerial platform}
    \label{fig:connect}
\end{figure}

{\colc To handling computation tasks, the onboard computer serves as the central hub, bridging hardware and software across the entire autonomous system. For localization, the onboard computer processes image data from the QR camera by USB protocol, extracting the relative position and orientation of the QR code. It also receives distance measurements between two UWB labels and their respective anchors by UART protocol, calculating the UAV's relative position and heading. In target detection, the onboard computer processes real-time images from the detection camera and extracts cargo information using a pre-trained CNN model. For path planning, it receives pose estimates of the target deck via Wi-Fi from the landing platform and generates search waypoints accordingly. In flight control, the onboard computer utilizes localization, perception, planning, and IMU data to generate control commands based on the state machine. These commands are transmitted via OSDK with UART protocol to DJI's low-level controller for execution. The state machine within the onboard computer determines the UAV's status by collecting data from various subsystems, while also controlling when to operate the motors by RS485 protocol to adhere the cargo.
Additionally, the interactions between various modules within the onboard computer are all based on the ROS system. This means that data read through protocols such as USB, UART, and others are converted to ROS messages for communication between the modules.}
For a visual representation of the onboard device connections centered around the edge computer, refer to Figure \ref{fig:connect}.

\section{Autonomous Sub-System}
The primary objective of this challenge is to transition unmanned autonomous systems from laboratory experiments to real-world missions. To achieve this, we have devised an autonomous flight architecture tailored for GNSS-denied environments, centered around a purpose-built aerial transportation platform. This architecture is propelled by a suite of mission-specific algorithms.
Subsequently, we delve into the detailed implementation of each sub-system in sequence.

\subsection{Coordinate System}
To establish a dynamic model of the air transportation platform, it is essential to define the coordinate system of the mission scene. Given that transportation missions take place at sea, where the environment is inherently dynamic, it necessitates the use of multiple coordinate systems and transformations. For clarity, we present the pertinent coordinate systems in Table \ref{tab:frame}.

\begin{table}[!htbp] 
\centering 
\vspace{5pt} 
\caption{The coordinate systems for different entities} 
\begin{tabular}{c|c|c} 
\hline
Frames & Axis & Description \\
\hline
$F_w$ & $\{{\bf x}_w, {\bf y}_w, {\bf z}_w\}$ & The inertial(world) frame\\
$F_a$ & $\{{\bf x}_a, {\bf y}_a, {\bf z}_a\}$ & The landing platform frame\\
$F_{a_u}$ & $\{{\bf x}_{a_u}, {\bf y}_{a_u}, {\bf z}_{a_u}\}$ & The UWB anchor frame\\
$F_{a_q}$ & $\{{\bf x}_{a_q}, {\bf y}_{a_q}, {\bf z}_{a_q}\}$ & The QR code panel frame\\
$F_b$ & $\{{\bf x}_b, {\bf y}_b, {\bf z}_b\}$ & The body frame of UAV\\
$F_{b_u}$ & $\{{\bf x}_{b_u}, {\bf y}_{b_u}, {\bf z}_{b_u}\}$ & The UWB label frame\\
$F_{b_q}$ & $\{{\bf x}_{b_q}, {\bf y}_{b_q}, {\bf z}_{b_q}\}$ & The QR camera frame\\
$F_{b_d}$ & $\{{\bf x}_{b_d}, {\bf y}_{b_d}, {\bf z}_{b_d}\}$ & The detection camera frame\\
$F_c$ & $\{{\bf x}_c, {\bf y}_c, {\bf z}_c\}$ & The cargo frame\\
$F_d$ & $\{{\bf d}_x, {\bf d}_y, {\bf d}_z\}$ & The cargo deck frame\\
\hline
\end{tabular}
\label{tab:frame} 
\end{table}

\begin{figure}[thpb]
    \centering
    \includegraphics[scale=0.5]{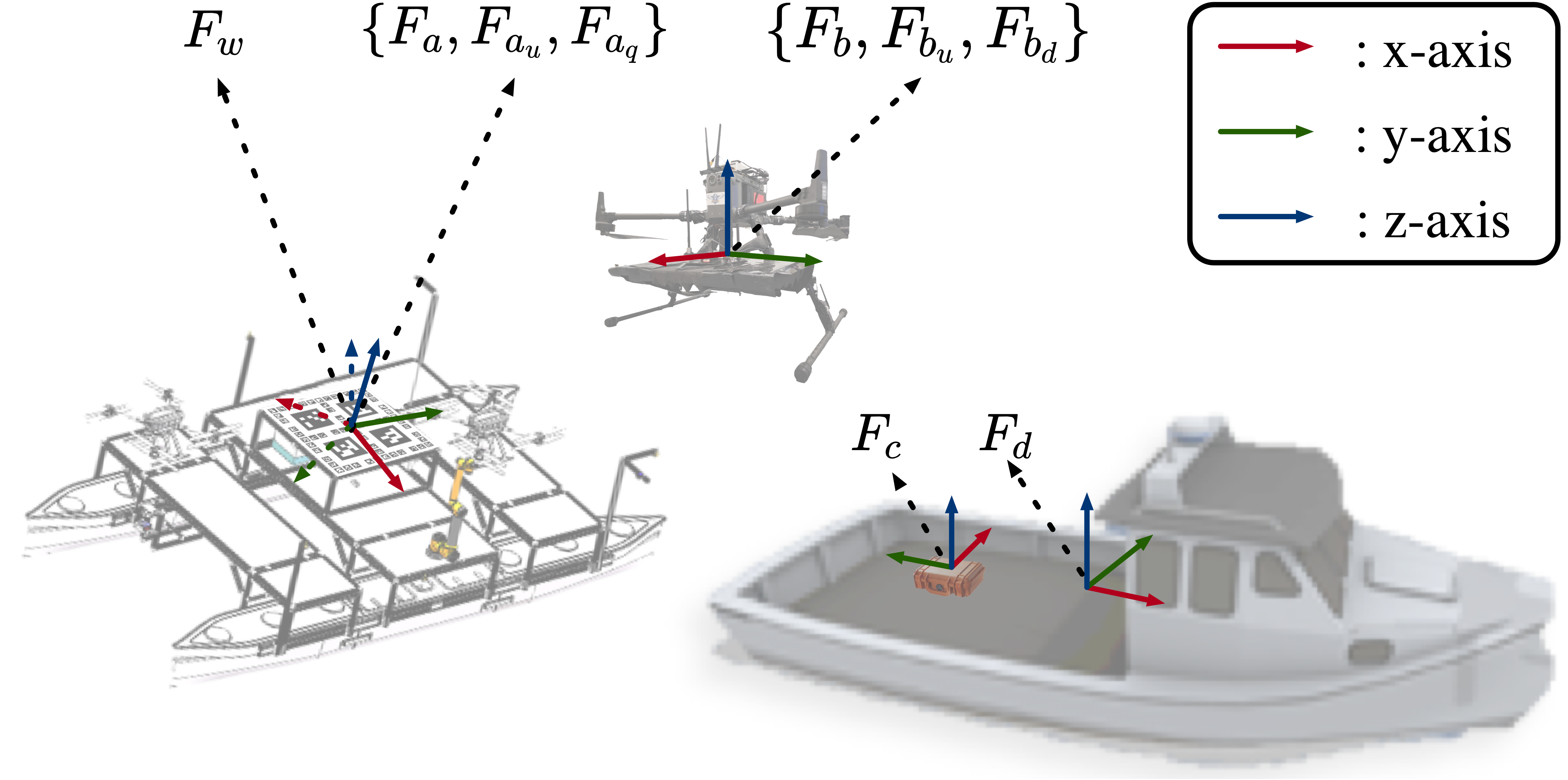}
    \caption{\colr The Schematic of the simplified coordinate system for the aerial transport system}
    \label{fig:frames}
\end{figure}

{\colm Considering that some coordinate transformations are fixed, even in dynamic environments, for the sake of clarity and simplicity in our discussion, we can align certain coordinate systems through a single transformation. This allows us to treat them as coincident, simplifying the analysis.}
Given the consistent attachment of UWB anchors and the QR code panel to the UAV's landing platform, we align frame $F_{a_u}$ and frame $F_{a_q}$ with frame $F_a$ via predetermined coordinate transformations.
{\colm Additionally, when the detection camera moves downward along with the carbon board, it will lose visibility of the cargo, thus no longer providing information. However, since the UAV will then proceed with its return process without needing cargo information, we simplify the description by assuming that the detection camera is fixed on the UAV and does not experience any relative movement.}
Subsequently, we align the frames of the detection camera $F_{b_d}$, the QR camera $F_{b_q}$, and the UWB label $F_{b_u}$ with the body frame $F_b$ in a similar manner.
{\colm Furthermore, noting that the landing platform moves on the sea surface, we treat the origin of the landing platform's coordinate system as the origin of the inertial coordinate system. However, due to the platform's oscillations, the inertial frame $F_w$ does not coincide with the landing platform's coordinate system $F_a$. This distinction is beneficial for subsequent algorithm design and analysis.}
The simplified coordinate system of the entire aerial transport system is depicted in Figure \ref{fig:frames}.

\subsection{Hybrid Localization Methods}\label{sec:HLM}
As discussed in the preceding section, the transport system relies on an artificial landmark-based localization approach, combining QR codes and UWB anchors, to navigate UAV in GNSS-denied environments.
Given the stringent positional accuracy required during UAV take-off and landing, the QR code localization method is employed.
Moreover, during search and return missions, the UAV may lose sight of the QR code within its field of view. In such instances, a UWB ranging-based localization scheme is activated.
Given that the roll angle $\phi$ and pitch angle $\theta$ of the UAV are obtained through the IMU, the primary objective is to calculate the UAV's position $p$ and yaw angle $\psi$ in frame $F_w$.

For QR code recognition, the open-source tools such as OpenCV\footnote{https://opencv.org} and ZBar\footnote{https://zbar.sourceforge.net} can be utilized for image processing, QR code recognition, and decoding. These tools allow for the extraction of each QR code's label and its position in frame $F_{a_q}$.
Additionally, the yaw angle of each QR code in the image can be determined by analyzing the pixel coordinates of the QR code's vertices. A detailed discussion on QR code identification and information extraction can be found in \cite{ref:HLI2015}.
It is important to note that the QR code board on the sea surface may not remain stationary. Combined with the fact that the UAV's attitude cannot be perfectly stable, this leads to geometric distortions of the QR code in the image.
Since the actual QR codes are square and their side lengths are known, geometric correction can be performed using a projection transformation. The specific methods for this correction are detailed in \cite{ref:RMC2003}.

{\colc To briefly outline the deployment of algorithms in this paper}, we assume the following facts are established and variables are known:
\begin{enumerate}
	\item Each QR code can be decoded into several variables, including the QR code label $i \in \mathcal{I}q$, the diagonal length $d_e^i$, {\colr and the coordinates $(q_x^i, q_y^i,0)$ in in frame $F_{a_q}$}.
	\item For the recognized QR code $i$ in the image plane, the QR code information includes the label $i$, the diagonal length $d_e^{i'}$, the  coordinates $(q_x^{i'}, q_y^{i'}, 0)$, and the yaw angle $\psi_{b_q}^{i'}$ in image frame $F_{b_q}'$.
	\item The image frame $F_{b_q}'$ coincides with the origin of frame $F_{b_q}$, {\colr differing only by a $\pi$ rotation in the yaw angle $\psi_{b_q}^{i}$ in $F_{b_q}$ for QR code $i$, that is $\psi_{b_q}^{i}=\psi_{b_q}^{i'}+\pi$.}
	\item With image geometry correction, the QR code in the image plane is parallel to the QR plane, meaning the ${\bf z}_{b_q}$-axis is parallel to the ${\bf z}_{a_q}$-axis.
	\item The focal length of the localization camera is $f_l$, and the distance from the camera pinhole to the QR plane is $f_g$.
	\item {\colr The attitude angles of the landing platform in frame $F_w$ are known, and frame $F_{a_q}$ is aligned with frame $F_{a}$. We assume that the Euler angles $\phi_{w}^{a}, \theta_{w}^{a}, \psi_{w}^{a}$ from frame $F_w$ to frame $F_{a_q}$ are also known.}
\end{enumerate}
\begin{figure}[thpb]
    \centering
    \includegraphics[scale=0.5]{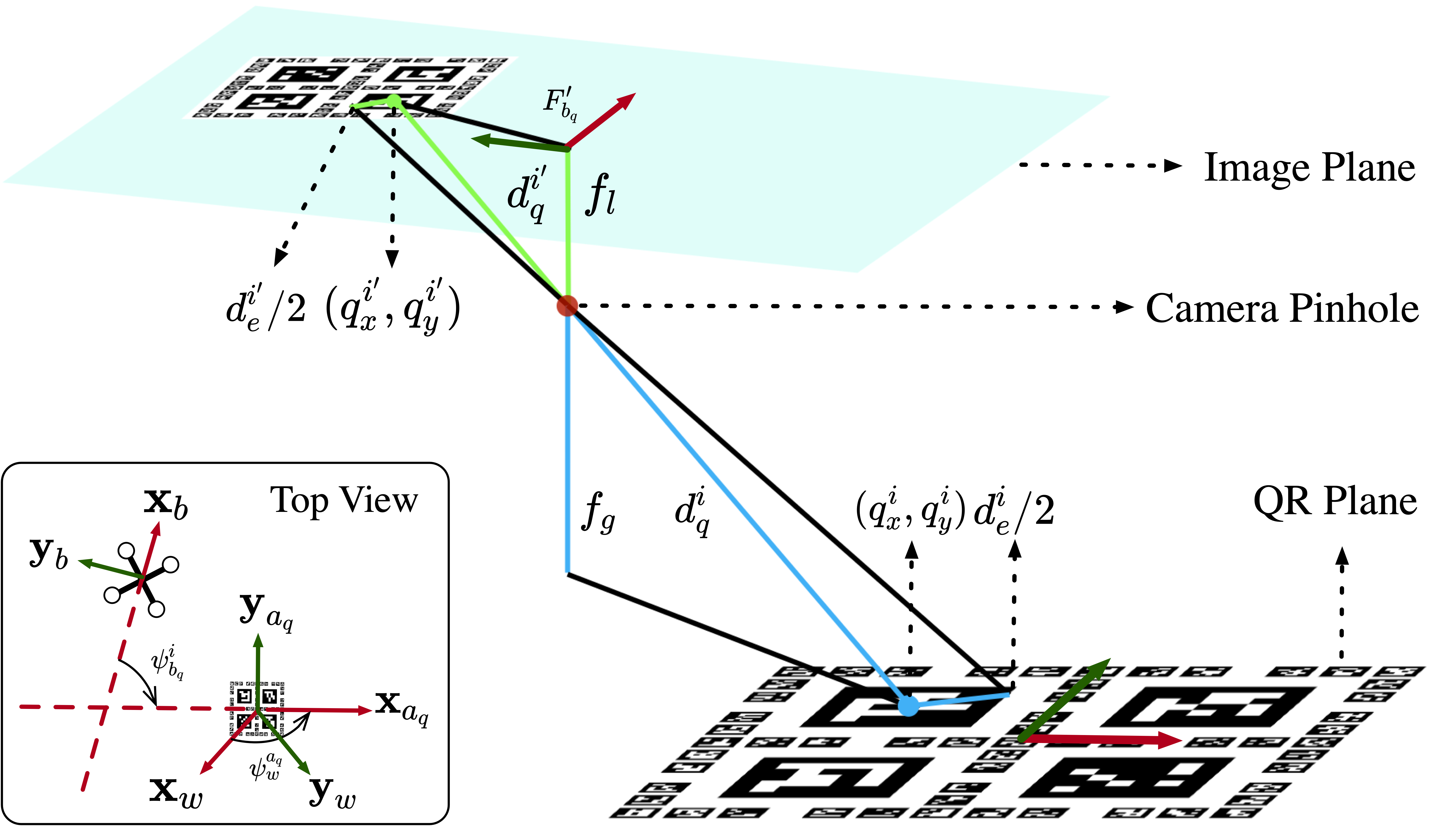}
    \caption{\colr The symbolic modeling in pose estimation using QR code}
    \label{fig:qr_position}
\end{figure}
To visually illustrate the aforementioned statement, we present the meaning of the variables in Figure \ref{fig:qr_position}.
Subsequently, we will use this information to calculate the position $p=[p_x,p_y,p_z]^T$ of the UAV in the frame $F_w$.
Based on the principle of similar triangles, the following relationship holds:
\begin{equation}\label{eq:flg}
	\frac{f_l}{f_g}=\frac{d_q^{i'}}{d_q^{i}}=\frac{d_e^{i'}}{d_e^{i}}
\end{equation}
where $d_q^{i}$ denotes the distance from the camera pinhole to the center of QR code $i$ in the real world, and $d_q^{i'}$ denotes the distance from the camera pinhole to the center of QR code $i$ in the image plane.
Then, we can obtain the ${\bf z}_{b_q}$-axis coordinate of QR code:
\begin{equation}\label{eq:zbi}
	z_{b_q}^i=-\frac{f_ld_e^i}{d_e^{i'}}-f_l.
\end{equation}
Similarly, we can determine the ${\bf l}_x$-axis and ${\bf l}_y$-coordinates of the center for the QR code $i$ in the frame $F_b$ as:
\begin{equation}\label{eq:xybi}
	x_{b_q}^i=\frac{q_x^{i'}d_e^i}{d_e^{i'}},\quad y_{b_q}^i=\frac{q_y^{i'}d_e^i}{d_e^{i'}}.
\end{equation}
Additionally, by equivalently transforming the yaw angle of the QR code in the image, derive the yaw angle of the UAV in the frame $F_w$ as: 
\begin{equation}
	{\colr \psi^i=\psi_w^{a_q}-{\psi_{b_q}^i}}
\end{equation}
where $\psi_w^{a_q}=\psi_w^{a}$.
A detailed discussion of the results for the above conversion process is presented \cite{ref:TAD2024}.
Through some trivial coordinate transformations, we can determine the position of the UAV in frame $F_w$ using the recognized QR code $i$:
\begin{equation}
	p^i=\left[\begin{array}{c}
		p_x^i\\
		p_y^i\\
		p_z^i
	\end{array}\right]=R_a^w\left[\begin{array}{c}
		q_x^i\\
		q_y^i\\
		0
	\end{array}\right]-R_b^{w}
	\left[\begin{array}{c}
		x_{b_q}^i\\
		y_{b_q}^i\\
		z_{b_q}^i
	\end{array}\right]
\end{equation}
where $R_a^w$ represents the rotation matrix determined by Euler angles $\phi_w^a,\theta_w^a,\psi_w^a$ and $R_a^w$ determined by $\phi,\theta,\psi$ with $\psi=\psi^i$.

For multiple QR codes recognized in the image, we obtain stable position and yaw angle by averaging their values:
\begin{equation}
	p=\sum_{i\in\mathcal{I}_q}\frac{1}{n_q}p^i,\quad \psi=\sum_{i\in\mathcal{I}_q}\frac{1}{n_q}\psi^i
\end{equation}
where $n_q$ denotes the number of recognized QR codes.
In conclusion, we have determined the position of the UAV in frame $F_w$ based on the QR code method.

Once the QR code is out of the camera's field of view, the transport platform will switch to the UWB-based localization method.
This method uses distance measurements to calculate the position of the UAV in frame $F_w$.
When the anchors and the distances between the labels are known, the position of the labels can be determined using trilateration measurements \cite{ref:FRI2004}. However, for smoother position information, we estimate the UAV's position based on the Extended Kalman Filter (EKF) method, which fuses IMU data. The EKF method for indoor UWB-based localization is discussed in detail in \cite{ref:DKI2020}.
This work focuses on highlighting the deployment and contribution of this algorithm in real marine environments.
The following facts and assumptions are considered for the UWB-based localization algorithm in this paper:
\begin{enumerate}
	\item The UWB anchors are deployed and fixed on the landing platform, which employs a common frame $F_{a_u}$. The rotation matrix $R_a^w$ from frame $F_{a_u}$ to frame $F_w$ are assumed to be known.
	\item In the frame $F_{a_u}$, the label positions at time $k$ are denoted as $u^i(k)=[u_x^i(k),u_y^i(k),u_z^i(k)]^T,i\in\{1,2\}$, and are symmetric along the center of the origin of $F_{b_u}$. The anchor positions are denoted as $\bar{u}^i=[\bar{u}_x^i,\bar{u}_y^i,\bar{u}_z^i]^T, i\in\{1,2,\dots,N_a\}$ where $N_a$ is the number of anchors.
	\item {\colc In the frame $F_{b}$, the coordinates of label 1 and label 2 are given as $u_b^1=[0,d/2,0]^T$ and $u_b^2=[0,-d/2,0]^T$ respectively.}
	\item The spatial distribution of anchors ensure observability for the label positions.
	\item IMU data includes the UAV acceleration $a^b(k)=[a_x^b(k),a_y^b(k),a_z^b(k)]^T$ in frame $F_b$ at time $k$, roll angle $\phi$ and pitch angle $\theta$ in frame $F_w$. {\colc The yaw angle $\psi$ from the IMU is unreliable due to magnetic interference.}
	\item {\colo The Euclidean distance between UWB labels $i$ and UWB anchors $j$ at time $k$ are denoted as $d_j^i(k)=\sqrt{(u_x^i(k)-\bar{u}_x^i)^2+(u_y^i(k)-\bar{u}_y^i)^2+(u_z^i(k)-\bar{u}_z^i)^2}.$}
\end{enumerate}

Considering the UAV in frame $F_{a_u}$, we define the state vector of label $i$ at time $k$ as $U_{a_u}^i(k)=[(u^{i}(k))^T,(\dot{u}^{i}(k))^T]^T, i\in\{1,2\}$.
Based on the uniform acceleration model of the UAV, the state equation with the sample time period $T$ can be expressed as:
\begin{equation}\label{eq:state}
	U_{a_u}^i(k+1)=AU_{a_u}^i(k)+Ba^u(k)+DW(k)
\end{equation}
where 
\begin{equation}
	A=\left[\begin{array}{cc}
		I_3 & TI_3\\
		0 & I_3\\
	\end{array}\right], 
	B=\left[\begin{array}{c}
		\frac{T^2}{2}I_3\\
		TI_3
	\end{array}\right],
	D=
	\left[\begin{array}{c}
		\frac{T^3}{6}I_3\\
		\frac{T^2}{2}I_3\\
	\end{array}\right]
\end{equation}
where $I_n$ denotes the $n$-dimensional identity matrix. Additionally, the vector $W(k)=[w_x(k),w_y(k),w_z(k)]^T$ denotes the process noise with zero mean and covariance matrix $Q$, and $a^u(k)=R_w^uR_b^w a^b(k)$ is the acceleration in frame $F_{a_u}$.
Here, $R_w^u$ is known, and $R_b^w$ can use the {\colc calibration value} at the beginning and will be subsequently estimated based on double labels.

Then, define the observation vector as $Z^i(k)=[d_1^i(k),d_2^i(k),\dots,d_{N_a}^i(k)]^T+V(k)$ where $N_a$ is the number of anchors and $V(k)=[v_1(k),v_2(k),\dots,v_{N_a}(k)]$ represents the observation noise vector with zero mean and covariance matrix $R$.
Since this is a nonlinear equation, the EKF method requires linearization, which involves calculating the Jacobian matrix:
\begin{equation}
	H(k)\triangleq
	\left[\begin{array}{cccc}
		\frac{\partial d^i_1(k)}{\partial u_x^i(k)} & \frac{\partial d^i_1(k)}{\partial u_y^i(k)}  & \frac{\partial d^i_1(k)}{\partial u_z^i(k)} & {\bf 0}_{1\times 3}\\
		\vdots & \vdots  & \vdots & \vdots\\
		\frac{\partial d^i_{N_a}(k)}{\partial u_x^i(k)} & \frac{\partial d^i_{N_a}(k)}{\partial u_y^i(k)}  & \frac{\partial d^i_{N_a}(k)}{\partial u_z^i(k)} & {\bf 0}_{1\times 3}
	\end{array}\right].
\end{equation}
The observation equation is then given by:
\begin{equation}\label{eq:obser}
	Z^i(k)=H(k)U_{a_u}^i(k)+V(k).
\end{equation}
Following the steps of the EKF algorithm \cite{ref:DKI2020}, the state estimate $U_{a_u}^i(k)$ of label $i$ in $F_{a_u}$ can be obtained.
Based on the distribution of labels, the corresponding state of the UAV in frame $F_w$ can be estimated as
\begin{equation}
	U_{a_u}(k)=\frac{U_{a_u}^1(k)+U_{a_u}^2(k)}{2}.
\end{equation}

{\colc Now, we have obtained the position of the labels in the frame $F_{a_u}$. Considering the issue of the landing platform's attitude oscillations, even when the UAV is hovering, the computed position will fluctuate due to these oscillations. The magnitude of this fluctuation will increase with the distance between the UAV and the landing platform. As a result, during the search or return process, if the control is based on this position, the drone may experience severe oscillations.
Therefore, we need to calculate the position in the inertial frame $F_w$, and then control the UAV based on the navigation points in frame $F_w$. This decouples the control from the platform's attitude oscillations.}
Then, since the origin of frame $F_{a_u}$ coincides with the inertial system $F_w$, the state estimate for UAV in frame $F_w$ can be given by
\begin{equation}
	U_{w}(k)=R_{a_u}^w U_{a_u}(k).
\end{equation}

{\colc Since the UAV's control is executed in its body frame, the position error calculated in the inertial frame needs to be further transformed into the UAV's body frame to serve as position error feedback. This transformation utilizes the UAV's Euler angles in the inertial frame, where the roll and pitch angles are obtained from the IMU, while the yaw angle $\psi$ is calculated using two fixed UWB labels mounted on the UAV.}
According to the definition of the rotation matrix $R_b^w$ and the vector $[\Delta x,\Delta y,\Delta z]^T=[u_x^1(k)-u_x^2(k),u_y^1(k)-u_y^2(k),u_z^1(k)-u_z^2(k)]^T$ constructed by label $1$ and label $2$, there is the relation:
\begin{equation}
	\left[\begin{array}{c}
		\Delta x \\ \Delta y \\ \Delta z
	\end{array}\right]=R_b^w{\colc (u_b^1-u_b^2)}=d\left[\begin{array}{c}
		\mathcal{S}_{\phi}\mathcal{S}_{\theta}\mathcal{C}_{\psi}-\mathcal{C}_{\phi}\mathcal{S}_{\psi} \\ \mathcal{S}_{\phi}\mathcal{S}_{\theta}\mathcal{S}_{\psi}+\mathcal{C}_{\phi}\mathcal{C}_{\psi} \\ \mathcal{S}_{\phi}\mathcal{C}_{\theta}
	\end{array}\right]
\end{equation}
where the shorthand notations ``$\mathcal{S}$" and ``$\mathcal{C}$" represent $\sin$ and $\cos$, respectively.
After several conversions, we calculate the yaw angle as:
\begin{equation}
 \psi=\left\{
 \begin{array}{l}
   {\rm arctan2}(-\Delta x,\Delta y), \quad {\rm if}\ \mathcal{S}_{\phi}\mathcal{S}_{\theta}=0,\\
   {\rm arctan2}(\rho_1,\rho_2) \quad {\rm otherwise}.
 \end{array}\right.
\end{equation}
where $\rho_1=\frac{\Delta y}{d}\mathcal{S}_{\phi}\mathcal{S}_{\theta}-\frac{\Delta x}{d}\mathcal{C}_{\phi}$ and $\rho_2=\frac{\Delta x}{d}\mathcal{S}_{\phi}\mathcal{S}_{\theta}+\frac{\Delta y}{d}\mathcal{C}_{\phi}$.

{\colc Now, we obtain the UAV's pose using both QR code-based and UWB-based localization methods. QR-based localization primarily operates during the takeoff and landing phases, and if QR code-based method becomes unavailable due to external factors, the system will automatically switch to UWB-based method.
Conversely, UWB-based localization is mainly employed during the search and return phases. If QR code-based method becomes available, the system will switch back to this method.
The QR code-based localization is prioritized due to its higher accuracy over the takeoff and landing area.
Additionally, to smooth out the transition between the two localization methods, the system applies mean filtering to the positioning data, thereby achieving a seamless switching process.}

\subsection{Target Detection}
Target detection performance, as a core metric of an autonomous transport system, decisively influences whether the UAV can accurately land on top of the cargo to execute the grasping maneuver. The core detection algorithm is based on the YOLO convolutional neural network (CNN) \cite{ref:JAP2022}.
The network was trained on data collected in simple environments at the lake and beach (see Figure \ref{fig:test_site}) to simulate the real race environment as much as possible. The output of the YOLO CNN is a set of candidate objects, each with its class, confidence, pixel position of the center, and bounding rectangle. The design ideas for the target detection method in this work are given below.

As the UAV moves, the image changes with it, leading to changes in the confidences of each candidate target. Typically, the target with the highest confidence is considered to be the transport cargo. However, when there are multiple objects on the deck, changes in confidence may cause the identification of the transport cargo to jump back and forth, disrupting the UAV's control stability.
To address this issue, we implement a wavegate strategy. When an object is stably recognized, meaning it has the highest confidence over multiple consecutive frames, the region near the target is cropped and used as the image input to the CNN. This cropped region dynamically changes with the position and size of the object's bounding rectangle to ensure that it contains only a single target. If the target is lost, the entire image is used again as the CNN input.

Similar to the principle of localization using the camera, as detailed in Eqs. (\ref{eq:flg}), (\ref{eq:zbi}), and (\ref{eq:xybi}), the position of the target at time $k$ in the frame $F_b$ can be calculated as $c_b=[x_{b_c},y_{b_c},z_{b_c}]$. The yaw angle of the cargo in frame $F_b$ is denoted as $\psi_b^c$.
Since the camera and the UAV are solidly connected, changes in the UAV's attitude will inevitably cause changes in the target's position in the image. This can lead to instability in UAV control during the final landing phase.
However, as the UAV approaches the target, the target position estimation error will continue to decrease.
To obtain stable target information, we smooth the target position using singular elimination and mean value filtering \cite{ref:POF1993}. Additionally, we estimate the velocity of the target, $\dot{c}_b$ by Kalman Filter \cite{ref:FAC1995}.

\subsection{Search Path Planing}
\begin{figure}[thpb]
    \centering
    \includegraphics[scale=0.5]{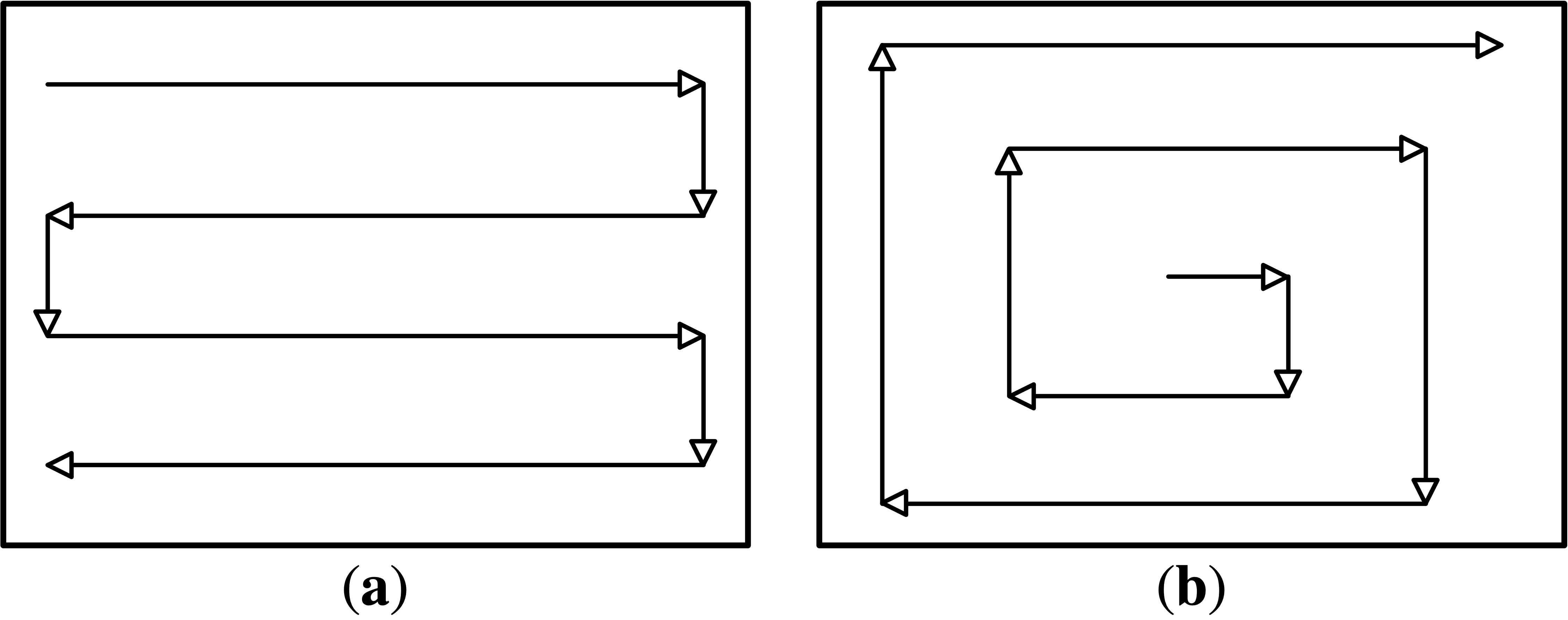}
    \caption{The simple coverage path schematic for a rectangular area: ({\bf a}) back-and-forth; ({\bf a}) spiral.}
    \label{fig:path}
\end{figure}

Since the location of the target cargo is unknown, the aerial transport system needs to autonomously generate a coverage path, addressing the Coverage Path Planning (CPP) problem.
Given the rectangular shape of the deck, a simple geometric pattern is sufficient to cover such areas. For simple convex polygonal areas, both back-and-forth and spiral patterns are considered effective CPP approaches in terms of energy efficiency \cite{ref:TSD2019}, which are demonstrated in Figure \ref{fig:path}.

{\colo In this study, we primarily employ a spiral pattern for coverage search, optimized for a rectangular search area (the deck). Given the simplified environment, we assume the UAV follows a constant-speed model, with energy efficiency measured by the minimum search distance. Considering that strong clutters from the marine surface affect the accuracy of UWB-based localization \cite{ref:LMI2008}, we start the search from the center of the deck. More detailed energy consumption models for UAV acceleration, deceleration, and turning can be found in \cite{ref:DJC2016}, with specific efficiency metrics further discussed in \cite{ref:TEI2023}.}

To achieve a specific coverage path, the deck should be modeled as a 2D grid. The dimensions of this grid depend on the coverage area of a single frame captured by the UAV's detection camera and the known side lengths of the deck. The side length of each cell can be calculated based on the field of view angle and height as follows:
\begin{equation}
	L=2z_{b_c}\cdot\max\{\tan(\frac{v_{\rm fov}}{2}),\tan(\frac{h_{\rm fov}}{2}\})
\end{equation}
where $v_{\rm fov}$ denotes vertical field of view and $h_{\rm fov}$ denotes the horizontal field of view.
Using Algorithm \ref{alg:path}, it is then possible to generate a set of spiral paths, which can be represented by the ordered list $[(0, 0), \dots, (x, y), \dots]$.
Additionally, assuming the 2D coordinates of the deck center $[d_x^i,d_y^i]^T$ and yaw angle $\psi_d$ in frame $F_w$ are known, the coordinates $[p^i_x,p^i_y]^T$ of the path point $(x,y)$ in frame $F_w$ can be calculated as:
\begin{equation}
	p^i_{ij}=L\left[\begin{array}{cc}
		\mathcal{C}_{\psi_d} & \mathcal{S}_{\psi_d}\\ -\mathcal{S}_{\psi_d} & \mathcal{C}_{\psi_d}
	\end{array}\right]
	\left[\begin{array}{c}
		x \\ y
	\end{array}\right]
	+\left[\begin{array}{c}
		d^i_x \\ d^i_y
	\end{array}\right].
\end{equation}

{\colc Note that the detection range of a single frame is a 16:9 aspect ratio rectangle. The UAV can adjust its yaw angle (rotation around the vertical axis) by 180 degrees to cover different directions from the same position. This adjustment allows the UAV to scan a larger area efficiently, particularly when operating with limited field of view.}
Additionally, while a higher flight altitude increases the UAV's detection range, it also reduces the pixel density of objects in the image, making target recognition more challenging. The UAV will perform a horizontal search based on planning algorithms, starting from a certain critical altitude, until the cargo is detected or the entire deck area is covered. If the UAV fails to locate the cargo after covering the entire deck, it will descend to regenerate the search path, likely because the cargo was not recognized in the initial images.

{\colc In the search process, we primarily avoid collisions by restricting altitude, that is the UAV remains above the target ship's cabin until the cargo is detected. Upon detecting the cargo, the UAV first hovers at a safe height directly above the cargo, then descends. Clearly, this approach lacks universal applicability. Considering additional sensors for obstacle avoidance is an essential task and a potential area for future work, aimed at enhancing operational safety in complex and dynamic environments, such as maritime operations.}

\begin{algorithm}
\caption{{\colo Path Planning for Search}}
\label{alg:path}
\begin{algorithmic}[1]
\REQUIRE $m,n$ \COMMENT{Size of the grid}
\ENSURE $path$ \COMMENT{List of nodes in spiral order}

\STATE Initialize a grid of size $m \times n$ with all values set to 0
\STATE Define $directions$ as $[(0, 1), (1, 0), (0, -1), (-1, 0)]$ \COMMENT{Right, Down, Left, Up}
\STATE Set $current\_direction$ to 0
\STATE Set $spiral\_radius$ as $1$
\STATE Initialize $x, y$ to $(0, 0)$
\STATE Initialize $path$ as $[(x, y)]$
\STATE Mark $grid[x][y]$ as visited (set to 1)

\FOR{$i = 1$ to ($mn-1$)}
    \STATE $new\_x \leftarrow x + directions[current\_direction][0]$
    \STATE $new\_y \leftarrow y + directions[current\_direction][1]$
    \IF{$|new\_x|\leq spiral\_radius + (m-n)$ \AND $|new\_y|\leq spiral\_radius$ \AND $grid[new\_x][new\_y]\neq0$}
        \STATE $x \leftarrow new\_x$, $y \leftarrow new\_y$
    \ELSE
        \STATE $current\_direction \leftarrow (current\_direction + 1) \mod 4$
        \STATE $x \leftarrow x + directions[current\_direction][0]$
        \STATE $y \leftarrow y + directions[current\_direction][1]$
    \ENDIF
    \STATE Append $(x, y)$ to $path$
    \STATE Mark $grid[x][y]$ as visited (set to 1)
\ENDFOR

\RETURN $path$
\end{algorithmic}
\end{algorithm}

\subsection{Flight Control}
The movement of the UAV is driven by the thrust and torque generated by its four propellers and is typically modeled using rigid body dynamics \cite{ref:TGI2010}.
The primary objective of the low-level controller is to stabilize the UAV by managing its attitude and thrust during flight.
The DJI M300 platform integrates this low-level controller and provides an interface for velocity control commands in either the inertial frame or the body frame. To address the issue of internal heading drift, we used the body frame interface to control the flight platform.

Considering the Front-Left-Up convention, the velocity control commands are divided into four channels: the x-direction, y-direction, z-direction, and yaw angle.
Since the planned path positions use coordinates in the inertial frame $F_w$, the control error needs to be transformed as:
\begin{equation}
	e_p^b(k)=R_w^b(k)[p^*_w(k)-p_w(k)]
\end{equation}
where $e_p^b(k)=[e_x^b(k),e_y^b(k),e_z^b(k)]^T$ denotes the position error in the UAV's body frame, $p^*_w(k)$ and $p_w(k)$ represent the path position and UAV position in frame $F_w$ at time $k$. The rotation matrix $R_w^b$ depends on the roll and pitch angles from the IMU and the yaw angle from hybrid localization.
For the vision-based landing process, the position of the cargo in frame $F_b$ at time $k$ is obtained directly, so the position error is defined as $e_p^b(k)=c_b(k)$.
Given that the target cargo is rectangular, the UAV's yaw angle must be adjusted to maximize the contact area between the UAV and the cargo during landing. The yaw angle error can be defined as:
\begin{equation}
	e_{\psi}^b(k)=\psi_b^c(k)+\frac{\pi}{2}.
\end{equation}

Note that changes in the UAV system model before and after carrying cargo, as well as uncertain perturbations in the marine environment, may lead to instability of its physical platform. Fortunately, the classical PID controller has less dependence on the exact system model and can handle nonlinear disturbances by tuning the parameters \cite{ref:FUM2022}. To enhance the transport system's reliability, the errors in the four channels are corrected for UAV velocity by the PID controller:
\begin{align}
	v_i^*(k)=&k_{p_i}e_i^b(k)+k_{i_i}[e_i^b(k)+e_i^b(k-1)+\dots]\nonumber\\
	&+k_{d_i}\frac{e_i^b(k)-e_i^b(k-1)}{T}
\end{align}
where $i\in\{x,y,z,\psi\}$ represents the channel index, and $T$ represents the velocity signal period.
During the landing process, this computational command further adds the velocity of the target for fast alignment as: $v_i^*(k)+\dot{c}_b$.

{\colc Additionally, the PID parameters $k_{p_i},k_{i_i},k_{d_i}$ are different and should be adjusted to meet the specific needs of each process (takeoff, search, landing and return). For example, rapid responsiveness is essential during takeoff to quickly move out of the danger zone, while high precision is required during landing. Detailed parameter adjustment strategies for each process are provided in Section \ref{sec:RD}.}

{\colc To ensure flight safety, the velocity command is subject to saturation: $\bar{v}_i^*(k) = \text{sat}(v_i^*(k), v_{\max})$, where $\bar{v}_i^*(k) = v_{\max}$ when $v_i^*(k)\geq v_{\max}$, $\bar{v}_i^*(k)= -v_{\max}$ when $v_i^*(k)\leq -v_{\max}$, and $\bar{v}_i^*(k)  =v_i^*(k)$ otherwise.
Furthermore, during extended deviations caused by external disturbances (e.g., ship movements or gust oscillations), integral wind-up may occur. To address this, an anti-windup method is implemented: when saturation of the velocity signal is detected, only the counteracting integral component is accumulated, while the reinforcing component is ignored. Mathematically, the integral term accumulates only when $v_i^*(k-1) \geq v_{\max}$ and $e_i^b(k) \leq 0$, or when $v_i^*(k-1) \leq -v_{\max}$ and $e_i^b(k) \geq 0$. This approach prevents prolonged control saturation.
}
Then, the DJI M300's low-level controller receives the velocities $\bar{v}_i^*(k)$ from different processes and responds by driving its four rotors.

\begin{figure}[thpb]
    \centering
    \includegraphics[scale=0.5]{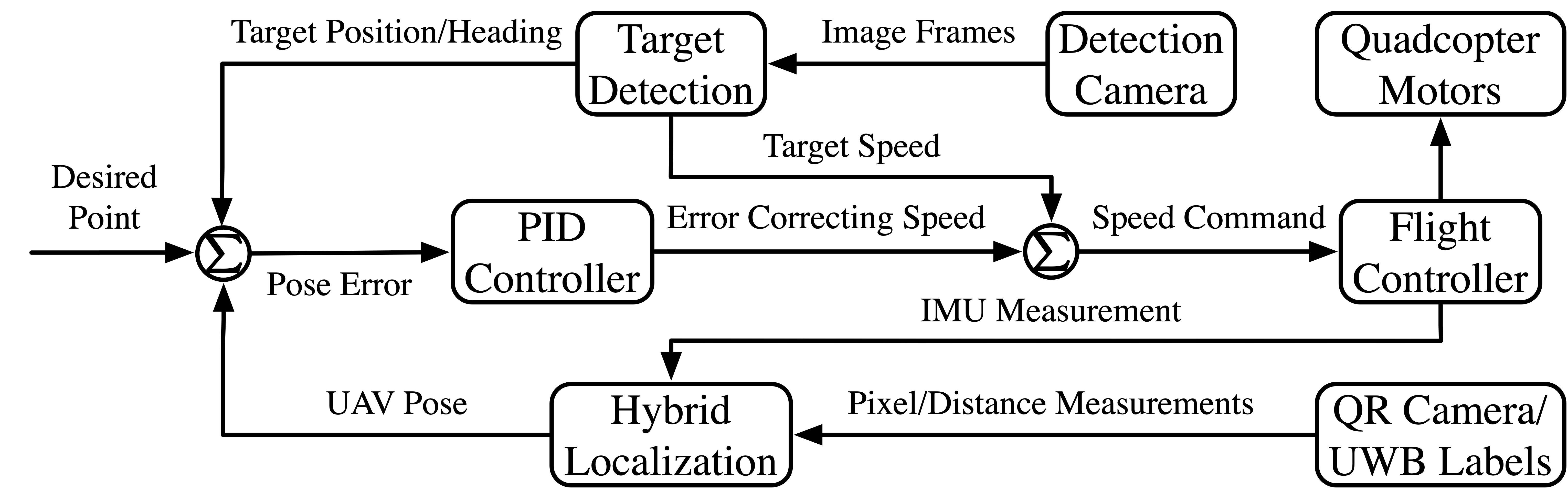}
    \caption{The high-level diagram of the system architecture}
    \label{fig:control}
\end{figure}

The control pipeline of the aerial transport system is designed for easy reconfiguration to accelerate development, and a high-level control diagram of its architecture is depicted in Figure \ref{fig:control}.
{\colo Specifically, during the search phase, the UAV determines its pose relative to the landing platform's coordinate frame using a UWB localization method integrated with IMU data. The high-level controller calculates the control error based on the current desired navigation point, and this error is used to generate velocity commands in body frame, which are then sent to the low-level controller for speed tracking. The high-level control strategy during the UAV's takeoff and landing phases on the platform is similar to the search phase, with the primary differences being the use of QR codes for localization and adjustments in control parameters. For descending above the cargo, the pose error feedback relies on recognition data of the cargo, and an additional velocity correction is applied based on the cargo's estimated movement to enhance tracking performance and landing precision. 
The low-level control layer utilizes a specific controller, such as PID or sliding mode control, integrated within the DJI platform. It generates the total thrust signal and the overall torque in the UAV's body frame, which are then transmitted to the actuators. This layer's purpose is to track the desired position or velocity provided by the high-level controller.

For a detailed explanation of the high-level and low-level control logic of quadrotor UAVs, including modeling and implementation, refer to \cite{ref:TFD2023}.}

\subsection{Cargo Attachment  Determination}
{\colb As shown in Figure 1, the UAV has a reflight mechanism after executing the transportation process to enhance the reliability of the transportation system. To balance the overall payload of the UAV, no additional sensors, such as cameras or LiDAR, were installed for judgment. Instead, the rotational speeds of the four rotors on the DJI platform are used for determination, which can be accessed via the OSDK.

Specifically, the thrust provided by the four rotors can be determined from their rotational speeds $\varpi_i, i=1,2,3,4$. When the UAV is hovering, the required thrust varies depending on the weight of its payload. Therefore, by comparing the difference in thrust before and after the transportation process, we can determine whether the UAV has successfully attached the cargo.
According to reference \cite{ref:MMI2012}, the total thrust of a quadrotor $T$ is proportional to the sum of the squares of the rotational speeds of the four rotors, as expressed by the following equation:
\begin{equation}
	T=c\left(\sum_{i=1}^4 \varpi _i^2 \right)
\end{equation}
where $c>0$ is a constant that can be determined from static thrust tests, but will not be used in this system.
Define the average thrust while the UAV is hovering before performing the adhesion operation as $\bar{T}$, and the average rotor speeds as $\bar{\varpi}_i$. After the adhesion operation, the average thrust is denoted as $\tilde{T}$, and the average rotor speeds as $\tilde{\varpi}_i$.
If $(\bar{T}_2 - \bar{T}_1) / c > \delta \bar{T}_1 / c$, i.e., $\sum_{i=1}^4 \bar{\varpi}_i^2 > \delta \sum_{i=1}^4 \tilde{\varpi}_i^2$, it is considered that the UAV has successfully attached the cargo and will return to landing platform. Otherwise, the UAV will reattempt flight to recapture the cargo.
It is worth noting that the thrust difference $(\bar{T}_2 - \bar{T}_1)$ can also be used to determine whether the attachment is successful. However, this threshold varies under different external conditions, such as indoor/outdoor environments or wind/no-wind scenarios, making it difficult to confirm reliably.

}

\section{Experimental Verification}
\subsection{System Performance Analysis}
\begin{figure}[thpb]
	\centering
    \includegraphics[width=0.6\linewidth]{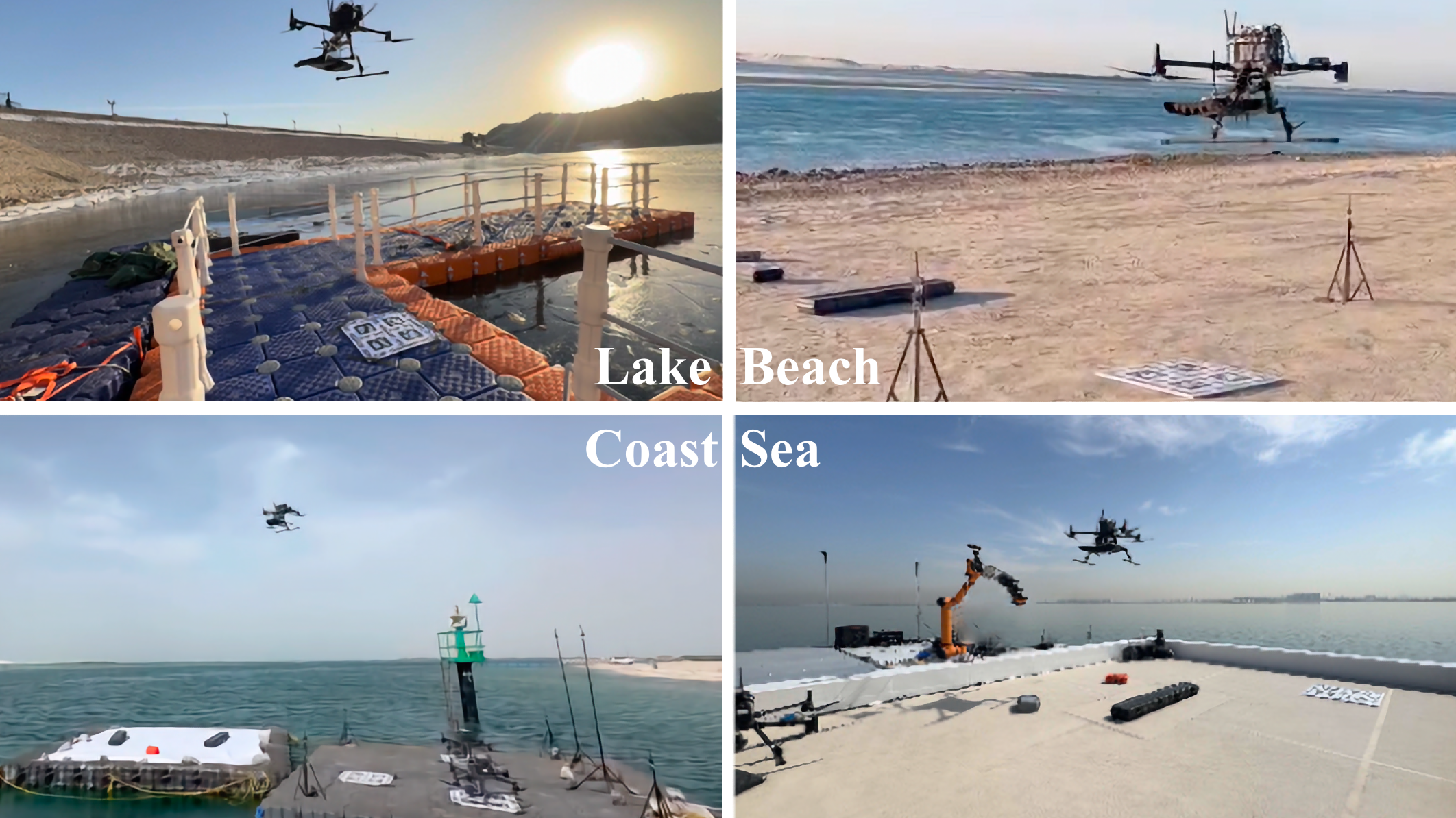}
	\caption{Four test sites simulate marine environments: in the top left, testing the grasping process on a lake; in the top right, assessing the platform's resilience to sea winds on a beach; in the bottom left, evaluating the landing accuracy of a mobile platform at the coast; and in the bottom right, conducting real competition tests at sea.}
	\label{fig:test_site}
\end{figure}

{\colm Two months prior to the final competition, we finalized the overall transport scheme and resolved all critical technical challenges}.
During this period, simulated marine environments were constructed both on a lake and on shore to train the object recognition model, optimize localization parameters and control gains, and size the grabbing structure, as depicted in Figure \ref{fig:test_site}. Extensive testing, including parameter and structural corrections, was conducted under varying light conditions, wind gusts, and random oscillations of the landing platform to refine a reliable onboard processing system suitable for a GNSS-denied environment.

\begin{figure}[thpb]
	\centering
    \includegraphics[width=0.8\linewidth]{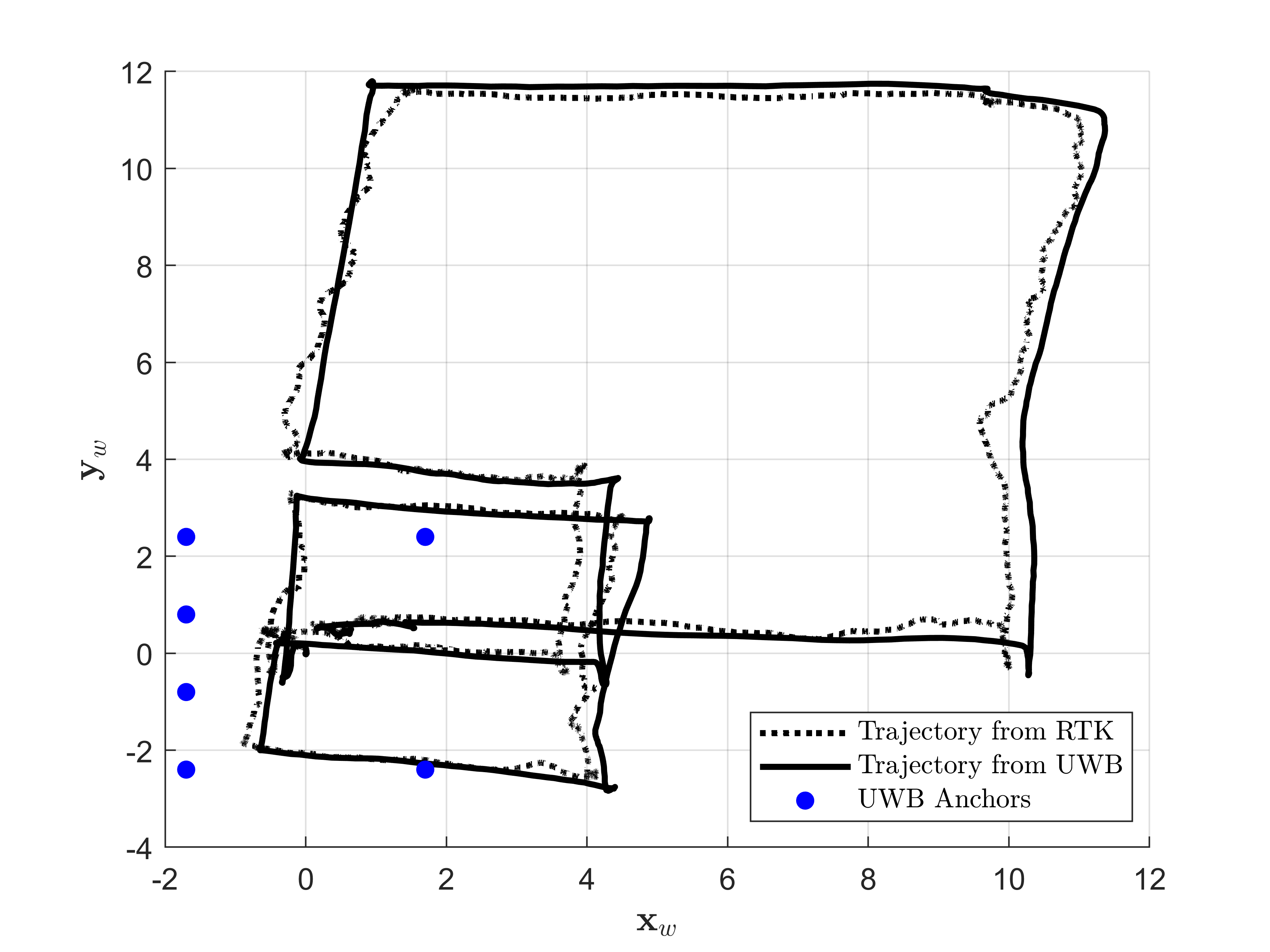}
	\caption{The estimation trajectories of UAV's position in frame $F_w$  based on RTK and UWB}
	\label{fig:loc_error}
\end{figure}

{\colr For the evaluation of system localization accuracy, we conducted an RTK-based assessment of the UWB localization method outside the competition field. In the experiment, six anchors were installed on UAV (landing platform), with the roll angle controlled within $\pm 8^{\circ}$ and the pitch angle within $\pm 10^{\circ}$. The UAV flied along a trajectory at an height of $5\rm{m}$ by Remote Controller (RC), as shown in Figure \ref{fig:loc_error}. The RTK-collected position data served as the benchmark, providing a horizontal accuracy of $1\rm{cm}$ and a vertical accuracy of $1.5\rm{cm}$. The specific accuracy of the UWB-based localization method, with RMSE as the metric, is shown in Table \ref{tab:loc_error_uwb}.
It can be observed that the UAV's estimation error is larger along the ${\bf x}_w$-axis, particularly when positioned outside the envelope of the anchors. In fact, this is determined by the spatial configuration of the spatial setup of anchors, similar to how GPS typically has lower accuracy in altitude estimation compared to horizontal positioning.
It is important to note that, as the most of trajectories fall within the outer bounds formed by all UWB anchors. However, because this phase corresponds to the UAV's search or return stage rather than the landing stage, the impact of these errors is comparatively minor.}

\begin{table}[!htbp] 
\centering 
\vspace{5pt} 
\caption{The RMSE of UWB-based localization method} 
\begin{tabular}{c|c|c|c} 
\hline
 & ${\bf x}_w$-axis & ${\bf y}_w$-axis & ${\bf z}_w$-axis\\
\hline
Hover above Origin  & 0.0113m & 0.0139m & 0.0212m\\
Moving by RC & 0.3010m & 0.1706m & 0.2280m\\
\hline
\end{tabular}
\label{tab:loc_error_uwb} 
\end{table}

{\colr Additionally, for the QR code-based localization method, the detection rate is nearly $100\%$ when the camera is within $5$m of the QR code. Given the significant variations in localization errors depending on the QR code's position in the image and the actual distance between the camera and the QR code, we provide a demonstration of performance by presenting the estimated error ranges at various distances (i.e., UAV heights) in Table \ref{tab:loc_error_qr}. Notably, if both the QR code and the QR camera remain stationary, the QR code-based method would not exhibit the large error fluctuations seen with UWB-based. Instead, its error tends to reflect a constant bias caused by installation deviations and image distortion.}

\begin{table}[!htbp] 
\centering 
\vspace{5pt} 
\caption{The estimation error range of the QR-based localization method at various distances.} 
\begin{tabular}{c|c|c|c|c|c} 
\hline
 & 0m$-$1m & 1m$-$2m & 2m$-$3m & 3m$-$4m & 4m$-$5m\\
\hline
Error Ranges  & 1cm$-$4cm & 3cm$-$9cm & 8cm$-$18cm & 16cm$-$30cm & 25cm$-$40cm\\
\hline
\end{tabular}
\label{tab:loc_error_qr} 
\end{table}

\begin{figure}[thpb]
	\centering
    \includegraphics[width=0.6\linewidth]{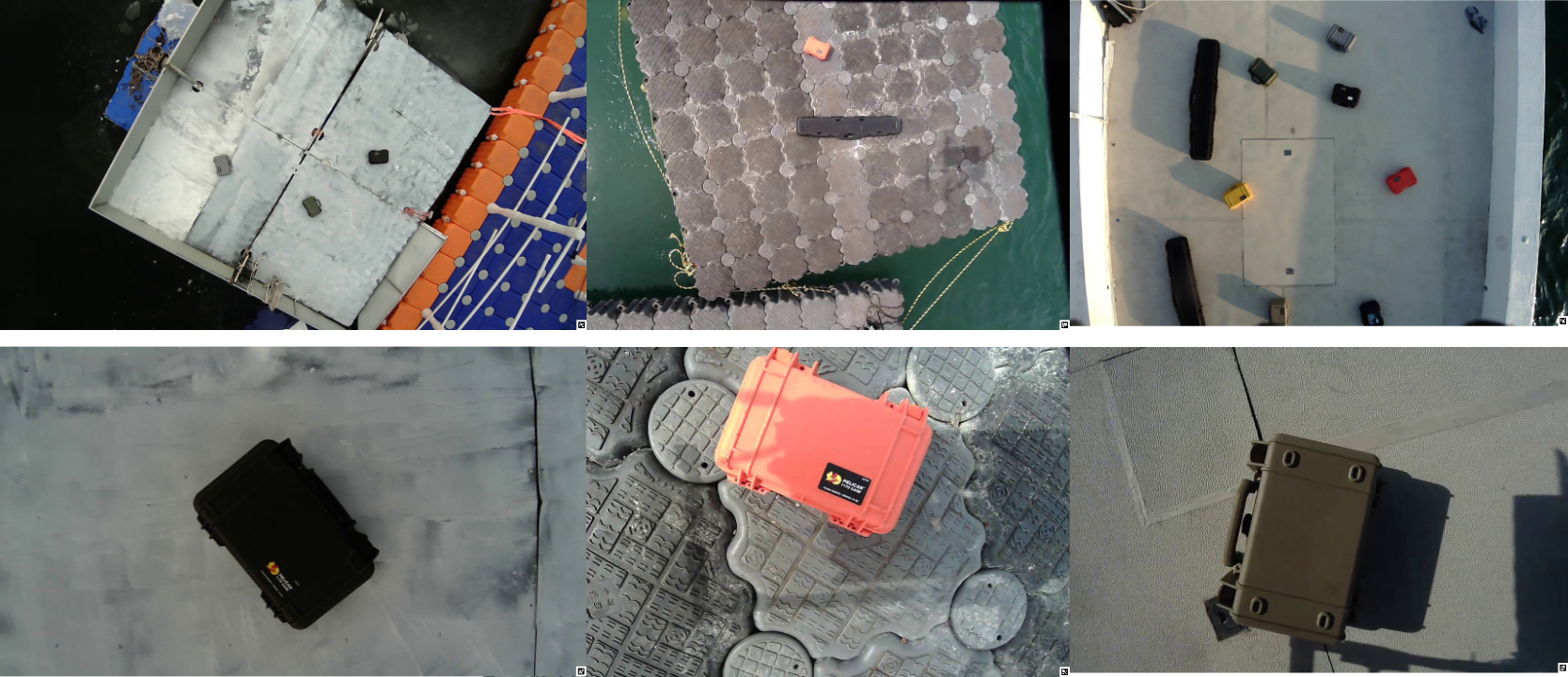}
	\caption{The illustration of the dataset for the cargo under different scenario conditions.}
	\label{fig:data_set}
\end{figure}

{\colr For the evaluation of detection performance, we collected over 6000 images across four different scenarios under varying lighting conditions. Of these, $80\%$ were used as the training set, all manually annotated (see Figure \ref{fig:data_set}), and the remaining $20\%$ were used as the test set. It should be noted that the dataset of the beach scene has been removed due to the significant difference in background from the racecourse. The model achieved an average precision (AP$\_$50) of approximately $92.4\%$ at an IoU threshold of 0.5. The inference time per frame on the Nvidia Jetson Orin NX was 12.7ms. However, in practical applications, the image acquisition rate is 30 FPS, and due to additional computation tasks, the actual output frame rate is approximately 21.3 FPS.}

{\colr In the evaluation of landing accuracy, the UAV employs visual servoing for landing control. During the final phase of landing, the UAV must stabilize 10cm above the cargo for alignment before proceeding to the next landing step. At this stage, the horizontal error must be less than 10cm, and the stabilization time is set to 3 seconds. Below 10cm in height, visual information becomes unavailable, prompting the UAV to perform a blind landing. Extensive testing under wind conditions ranging from level 0 to level 5 shows that the horizontal error resulting from blind landing does not exceed 5cm. In comprehensive transport tests conducted outside the competition setting, the UAV demonstrated over a $90\%$ probability of landing within 15cm of the cargo center. For the specific cargo and customized carbon board (supporting adhesive tape), as detailed in the following section, an effective contact area of $11\rm{cm}\times 15\rm{cm}$ can be achieved at this landing precision, assuming negligible relative heading angle error.}

{\colr In the evaluation of system adhesion performance, the ability to effectively attach cargo largely determines the success of the transportation task, which is also the motivation for implementing a reflight mechanism. In fact, the effective adhesion area and the humidity of the weather significantly affect the performance of the adhesive. To ensure redundancy for the final competition, we added an additional weight of 300g to the designated briefcase for testing. By manually controlling the UAV to perform maximum attitude angle oscillations, it is typically necessary to maintain an effective contact area of at least 100 square centimeters to ensure that the cargo does not fall. This test was conducted in December in Beijing, with an average humidity of $51\%$\footnote{https://www.climate.top/china/beijing/humidity}. Under normal transport conditions, excluding hardware failures and with adhesive tape replaced after four attachment attempts, the success rate of securing the cargo reaches $80\%$, even in challenging ocean environments affected by wind and waves. However, during January in Abu Dhabi, the reliability of adhesion was indeed reduced, with the local average humidity around $69\%$\footnote{https://www.climate.top/uae/abu-dhabi/humidity}. Extensive testing under equivalent conditions yielded a successful attachment rate of $70\%$. Additionally, we conducted simulated experiments along the shoreline, where dust significantly reduced the attachment rate to $50\%$. However, the attachment success can be improved by increasing the frequency of adhesive tape replacements.}

\subsection{Final Experimental Environment}
{\colr The final competition took place in a marine area near Yas Island, Abu Dhabi, covering approximately 10 square kilometers. The target vessel could be located anywhere within this area. In this study, we assume that the UAV's landing platform has already approached the vessel carrying the cargo, with a minimum separation of less than 3 meters. Due to ocean currents, neither the USV nor the target vessel could remain stationary relative to each other. Additionally, the exact location of the cargo was unknown. During the competition, participants were prohibited from intervening from the shore. The system operated specifically between 3:00 AM and 5:00 AM UAE time on February 5, 2024, during MBZIRC2024 final competition it was the only system to successfully complete the cargo transport task. A highlight video of the final competition can be viewed at the link\footnote{https://youtu.be/9hTAUkCvsy8}.}

\begin{figure}[thpb]
	\centering
    \includegraphics[width=0.6\linewidth]{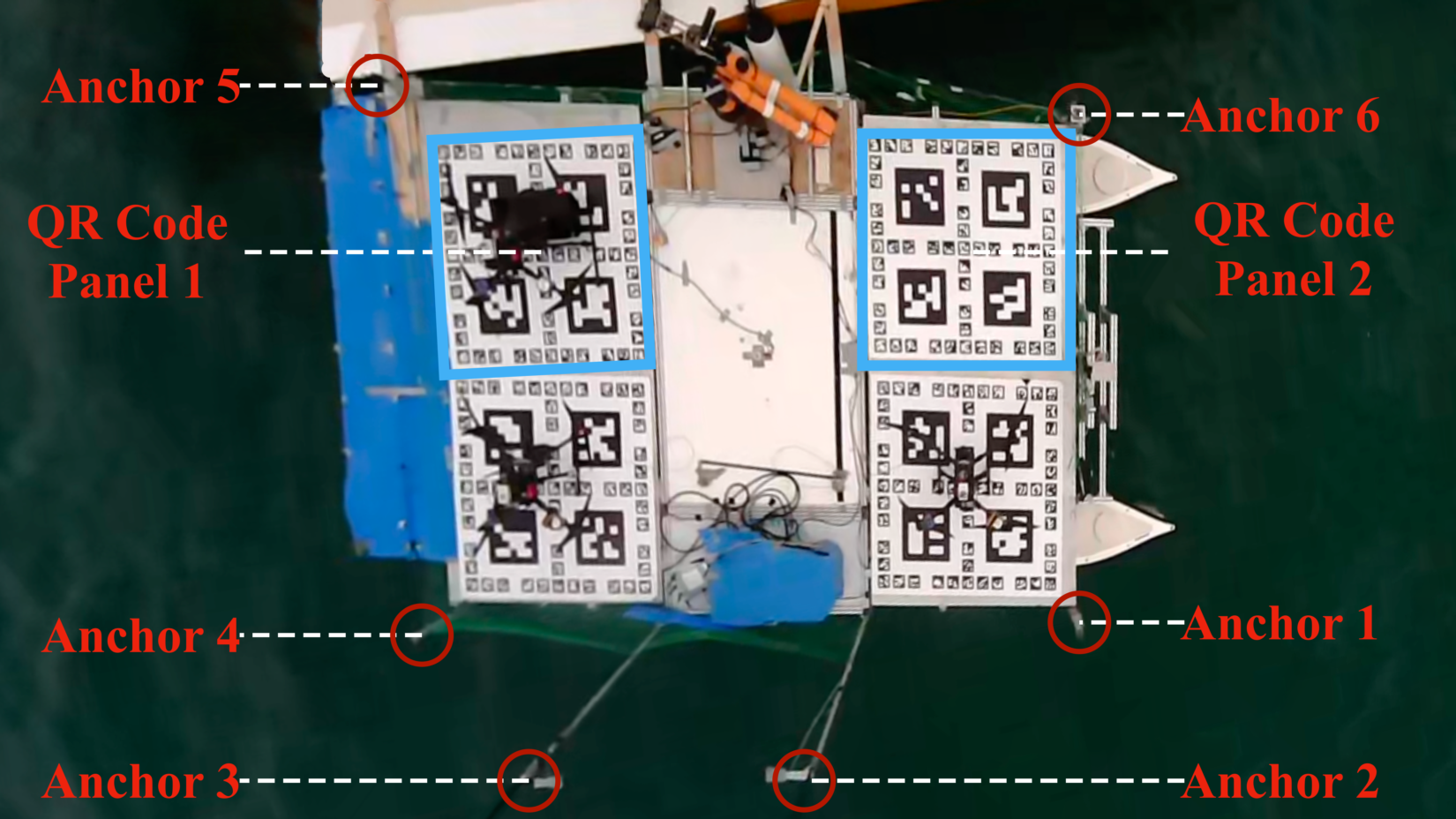}
	\caption{The landing and takeoff area for transport UAVs with QR code panels and UWB anchors.}
	\label{fig:deck}
\end{figure}

\begin{figure}[thpb]
	\centering
    \includegraphics[width=0.6\linewidth]{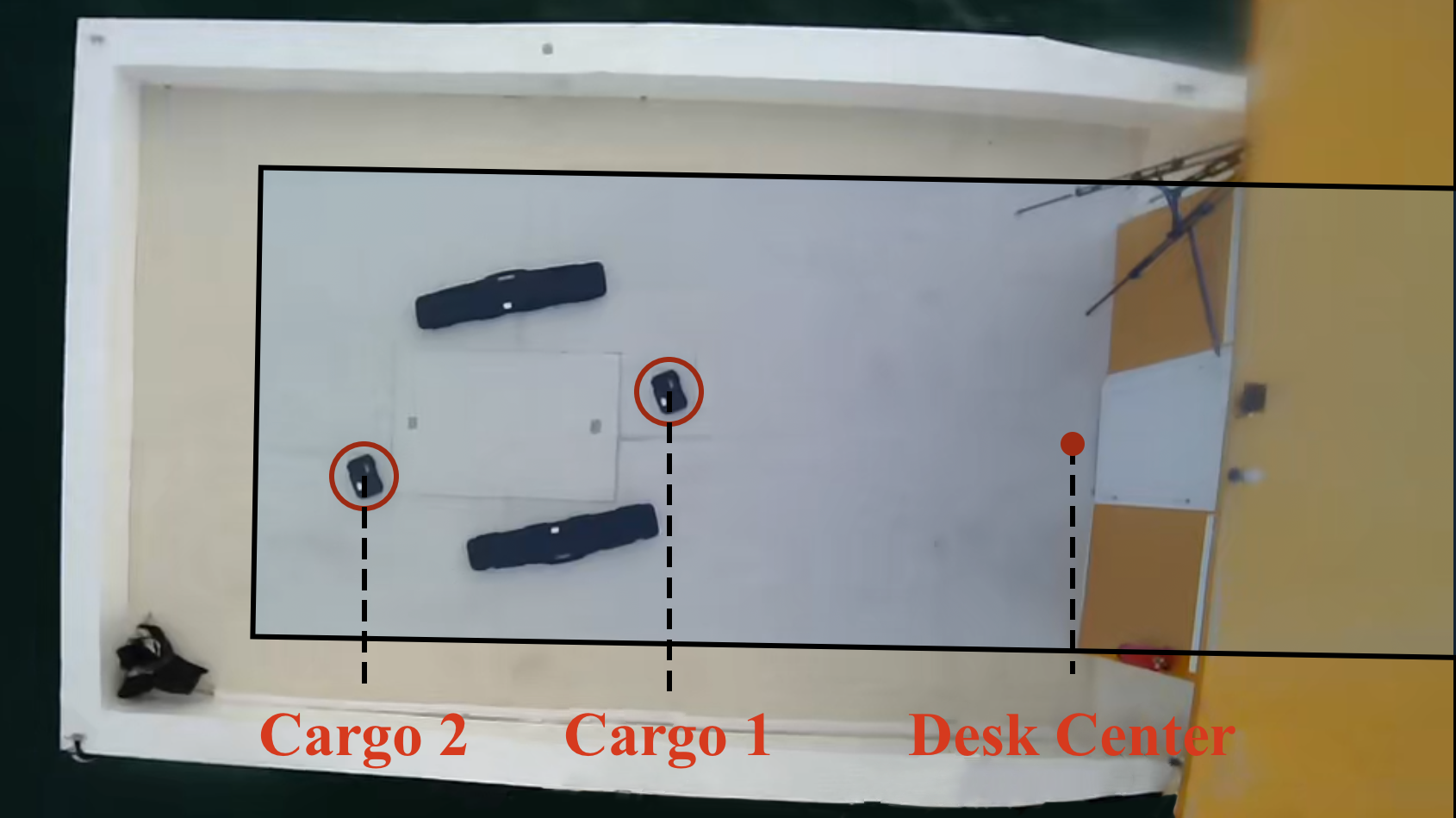}
	\caption{The deck of the target vessel on which the target cargoes are placed.}
	\label{fig:deck2}
\end{figure}

On the landing platform, measuring $3.5\rm{m}\times4.8\rm{m}$, four UAVs were stationed. Two of these UAVs carried two cargoes each on the target vessel's deck, while the remaining two UAVs were designated for other tasks not covered in this article. Each UAV was allocated a landing and takeoff area of $1.5\rm{m}\times1.5\rm{m}$ due to deck size limitations.
{\colr With the origin set at the center of the landing platform coordinate frame $F_a$, the positions of the two UAVs are $[1\rm{m}, 2\rm{m},0\rm{m}]$ and $[1\rm{m}, -2\rm{m}, 0\rm{m}]$, respectively.
Two matte boards with 142 different QR codes were fixed on the landing areas with same positions of two UAVs, and six UWB anchors were strategically positioned around the deck with positions $[\pm 1.7\rm{m},\pm 2.4\rm{m},0.2\rm{m}],[-1.7\rm{m},\pm 0.8\rm{m},3.7\rm{m}]$, as illustrated in Figure \ref{fig:deck}.}
The outer package of the target vessel measured $12\rm{m}\times6\rm{m}\times2\rm{m}$, with two cargoes randomly placed on its $4\rm{m}\times4\rm{m}$ deck, as shown in Figure \ref{fig:deck2}.
{\colr The cargo to be transported measures $30\rm{m}\times22\rm{m}\times10\rm{m}$ and has a net weight of 0.89 kg. However, the organizers may place objects inside the box, with the assurance that the total weight will not exceed $1$ kg. }
The color and load of the cargoes were randomized and specified by the organization.
{\colr To provide adequate contact area for the cargo, the carbon board in the adhesion mechanism was designed with size with $30\rm{cm}\times 50\rm{cm}$.}

On the day of the competition, the weather was cloudy with an average humidity of $56\%$. A northwest wind of 5 was blowing, with gusts reaching up to $12\rm{m/s}$ at the seaside. These conditions presented significant challenges for completing the object handling tasks in the harsh environment. However, they also served to verify the reliability of the aerial transport system to a greater extent.

\subsection{Competition Results Discussion}\label{sec:RD}
\begin{figure}[thpb]
	\centering
    \includegraphics[width=0.6\linewidth]{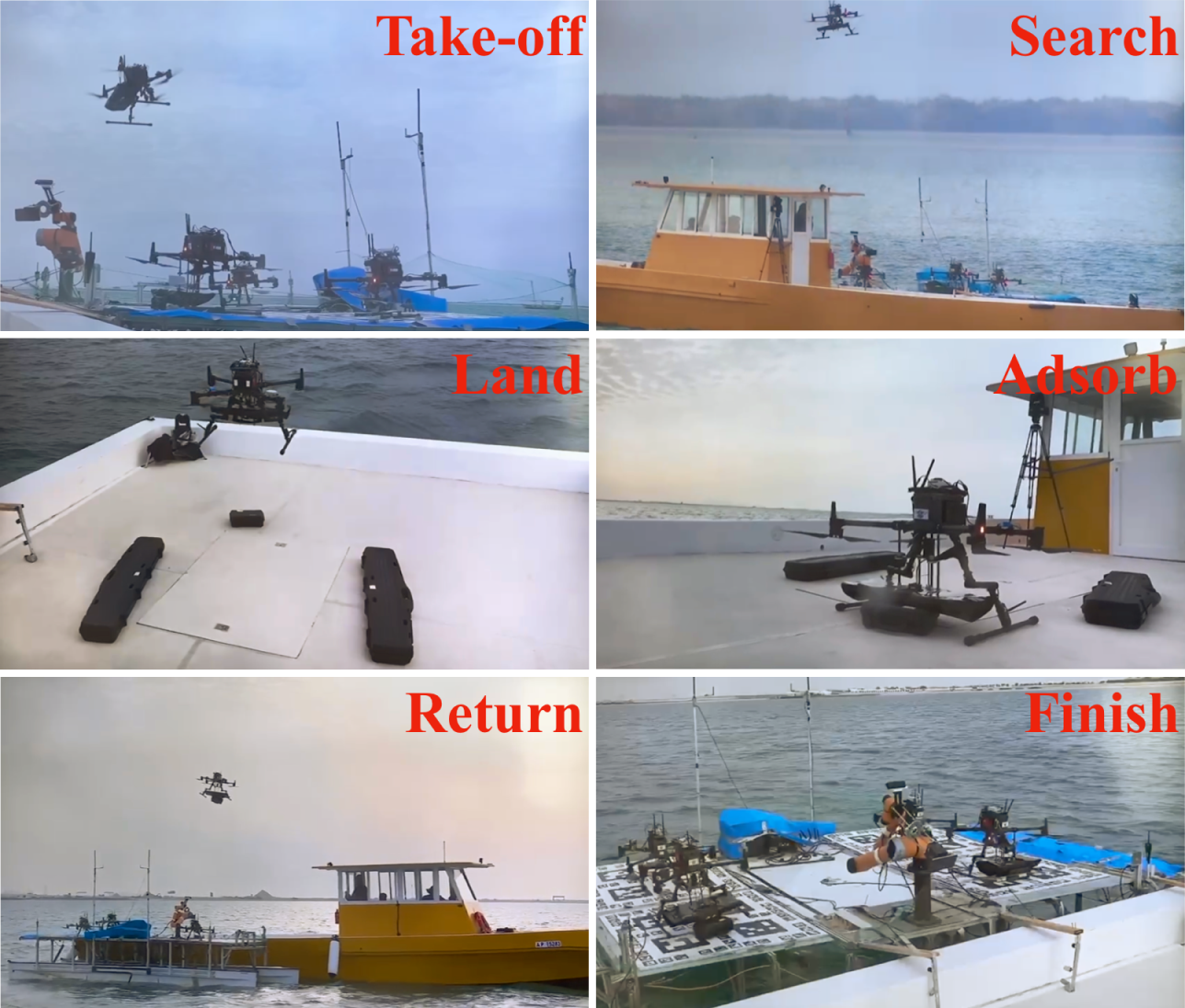}
	\caption{The demonstration of each process of transportation for the UAV in competition.}
	\label{fig:process}
\end{figure}

In the final competition, two UAVs of identical design, designated as UAV1 and UAV2, sequentially transported objects from the deck back to the landing platform. The overall transport process for both UAVs consisted of take-off, search, landing, manipulation, and return.
{\colo UAV 2 takes off 60 seconds after UAV 1 completes its mission, meaning that the tasks are executed sequentially without any overlap in the time dimension.}
A demonstration of the state of the UAV at each stage is depicted in Figure \ref{fig:process}. UAV1 and UAV2 took 229.87s and 158.50s, respectively, from take-off to transporting the target to the landing platform. The time spent on each process is detailed in Table \ref{tab:time}.

\begin{table}[!htbp] 
\centering 
\vspace{5pt} 
\caption{Summary for time spent at each process of transportation} 
\begin{tabular}{c|c|c|c|c|c} 
\hline
 & Take-off & Search & Land & Adsorb & Return\\
\hline
UAV1 & 9.56s & 15.52s & 34.84 & 19.58s & 79.20s\\
UAV2 & 9.13s & 35.00s & 41.22s & 35.60s & 108.92s\\
\hline
\end{tabular}
\label{tab:time} 
\end{table}

\begin{figure}[thpb]
	\centering
    \includegraphics[width=0.8\linewidth]{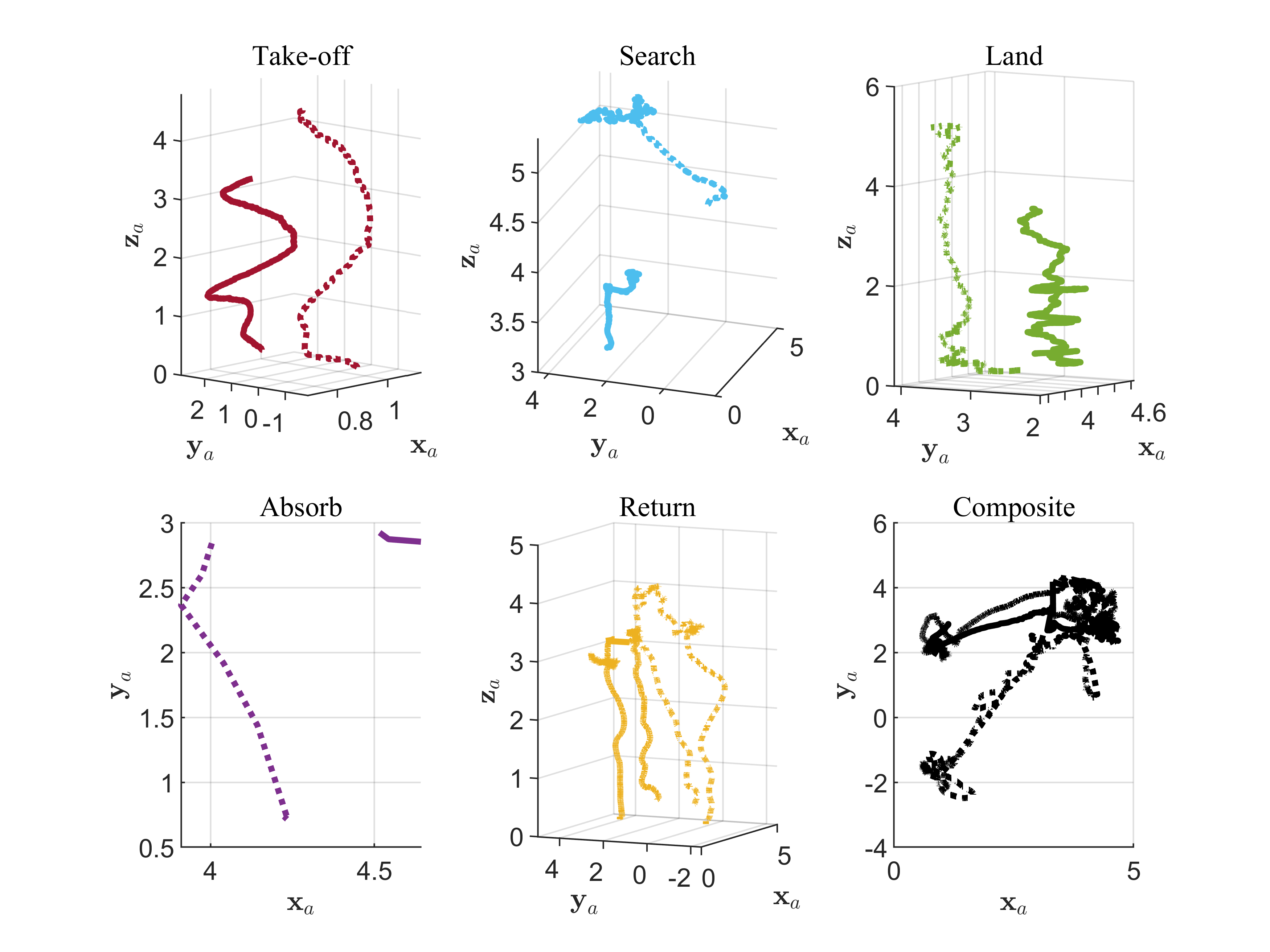}
	\caption{\colr The spatial trajectories of UAV1 and UAV2 in each process of transportation.}
	\label{fig:traj}
\end{figure}

{\colr Due to the GNSS-denied environment, reference trajectory data for the UAVs during transport operations is unavailable. To provide a visual analysis of the process, we generated estimated trajectories for UAV1 and UAV2 within coordinate frame $F_a$ based on hybrid localization data, as shown in Figure \ref{fig:traj}.
In the figure, the solid line represents the estimated trajectory of UAV1, while the dashed line represents UAV2. Significant oscillations are observed in the estimated trajectories, partly due to attitude oscillations of the landing platform.
During the takeoff phase, both UAV1 and UAV2 quickly reached the designated altitude. The positional error along the ${\bf y}_a$-axis is notably larger than along the ${\bf x}_a$-axis due to crosswinds parallel to the ${\bf y}_a$-axis. This outcome was anticipated, as the heading of the landing platform, without external thrust, naturally aligns with the ocean current.
In the search phase, the first waypoint for both UAVs was the center of the target vessel's deck. Fortunately, the estimated position of the landing platform relative to the deck was accurate, and the cargo appeared within the field of view, allowing the UAVs to proceed directly to the next phase upon cargo recognition.
Following this, the UAVs descended over the cargo based on visual detection information. It can be observed that UAV1 exhibited more pronounced oscillations than UAV2 at lower altitudes. This difference was due to UAV1's descent position being closer to the ship's cabin, where UWB signals experienced interference or occlusion, resulting in larger estimation deviations.
During the adhesion phase, UAV2's estimated trajectory showed considerable shifts, despite it being stationary on the target deck. This deviation occurred because the landing platform had drifted under the influence of ocean currents.
In the return phase, both UAVs required an extended duration, with noticeable trajectory oscillations. This was primarily due to the added payload, slightly affecting their overall speed tracking performance.
}

\begin{figure}[thpb]
	\centering
    \includegraphics[width=0.8\linewidth]{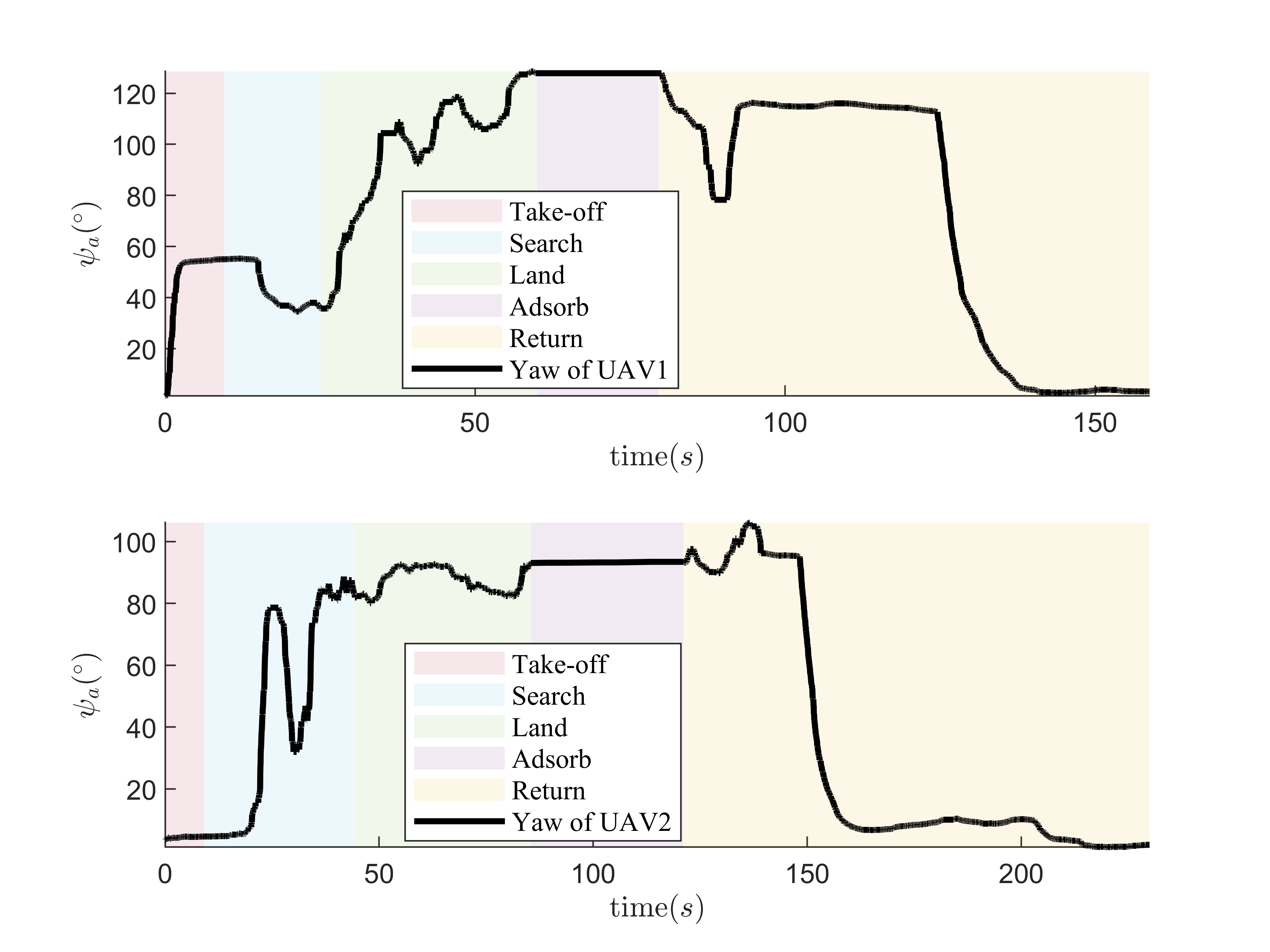}
	\caption{The yaw angle curves of UAV1 and UAV2 in each process of transportation.}
	\label{fig:yaw}
\end{figure}

{\colr Similarly, due to the lack of a reference, Figure \ref{fig:yaw} presents the curves of estimated relative yaw angle $\psi_a$ for UAV1 and UAV2 in frame $F_a$.
During the takeoff phase, UAV1 experiences rapid angular changes. This is due to magnetic interference that triggered a forced yaw correction by the DJI M300 low-level controller. However, it's worth noting that, despite the calibration, the angle remains imprecise in the inertial frame, due to the GNSS module's removal, resulting in an inaccurate correction.
In the search phase, yaw angle adjustments mainly occur as the UAV rotates about the vertical axis to extend its detection range. During descent over the cargo, the UAV rotates to align with the cargo, optimizing adhesive contact area for the waterproof seal. Finally, in the return phase, yaw adjustments ensure alignment with the landing platform, facilitating descent with the initial attitude.}

\begin{figure}[thpb]
	\centering
    \includegraphics[width=0.8\linewidth]{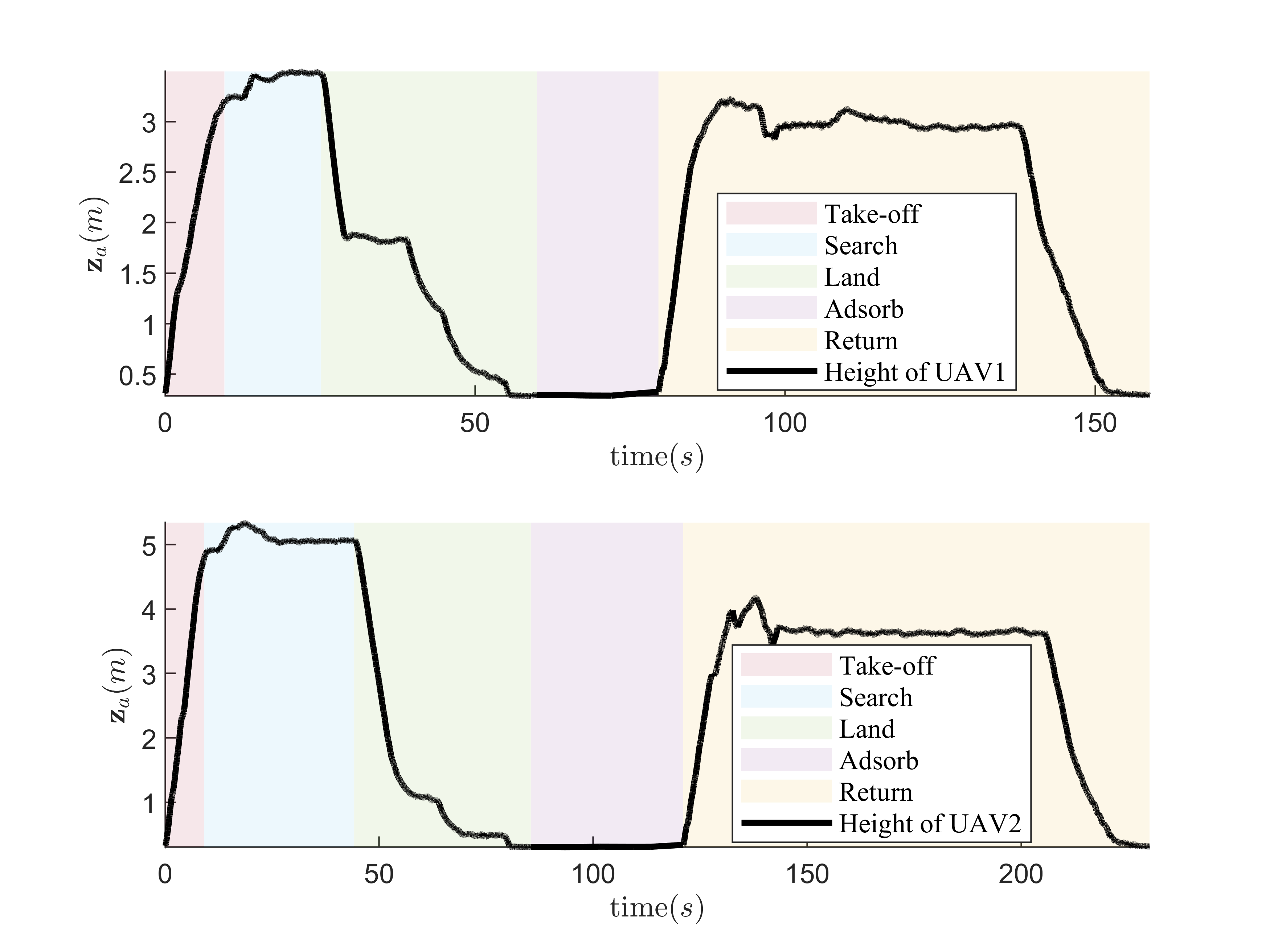}
	\caption{\colr The height curves of UAV1 and UAV2 in each process of transportation.}
	\label{fig:height}
\end{figure}

The height curves of the two UAVs during transportation are depicted in Figure \ref{fig:height}.
It is noticeable that there were two jumps in both the search and return phases, primarily attributed to the switching of localization methods{\colc (see Section \ref{sec:HLM}).}
During the landing phase, the UAVs performed two height hovers. The first hover was for adjusting the yaw angle, while the second was for executing a blind landing. This blind landing occurred after precisely aligning the cargo horizontally, demonstrating high precision towards the end of the landing process.

\begin{figure}[thpb]
	\centering
    \includegraphics[width=0.8\linewidth]{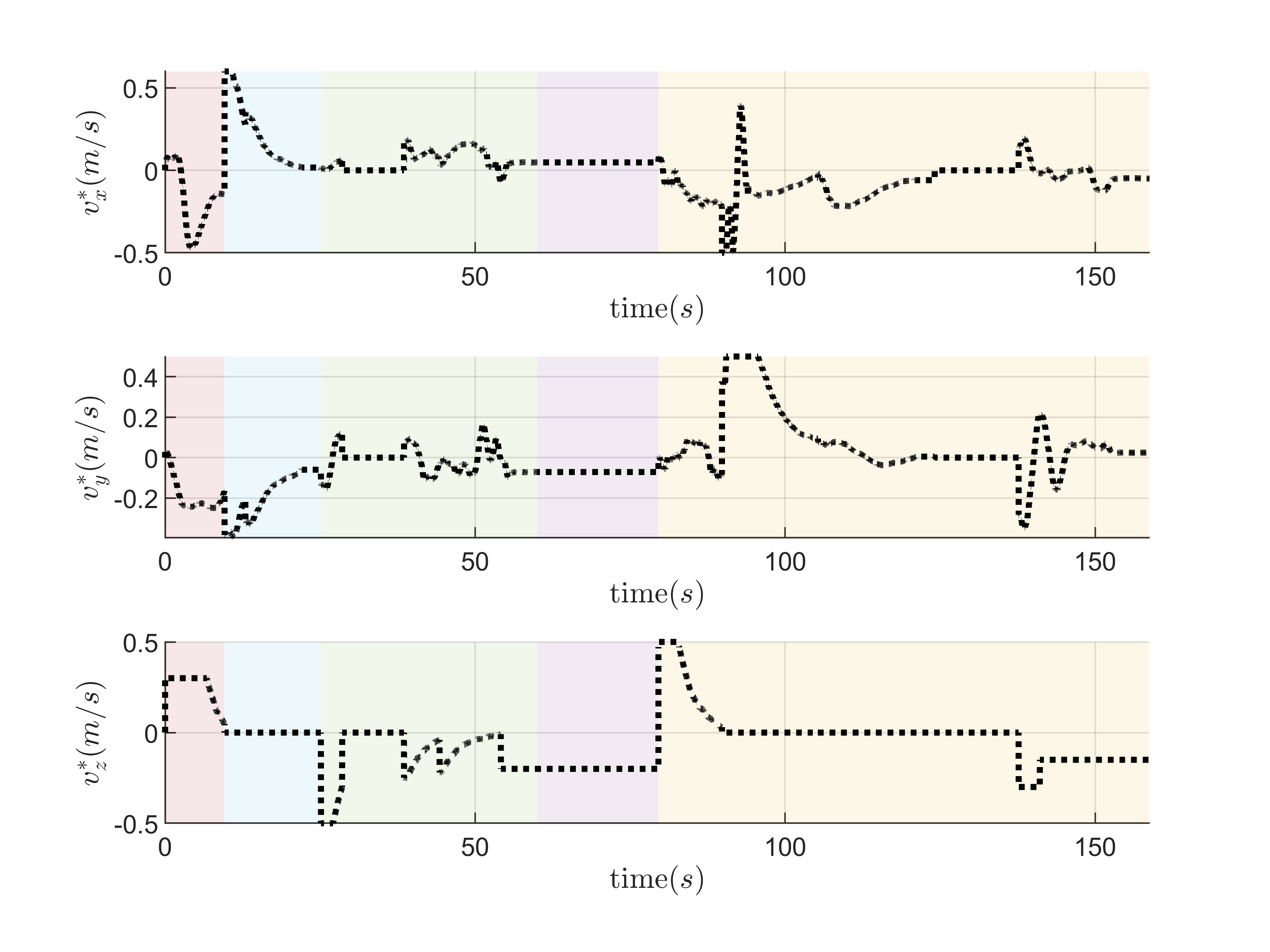}
	\caption{The three-axis velocity command curves of the UAV1 in each process of transportation.}
	\label{fig:vel_1}
\end{figure}

\begin{figure}[thpb]
	\centering
    \includegraphics[width=0.8\linewidth]{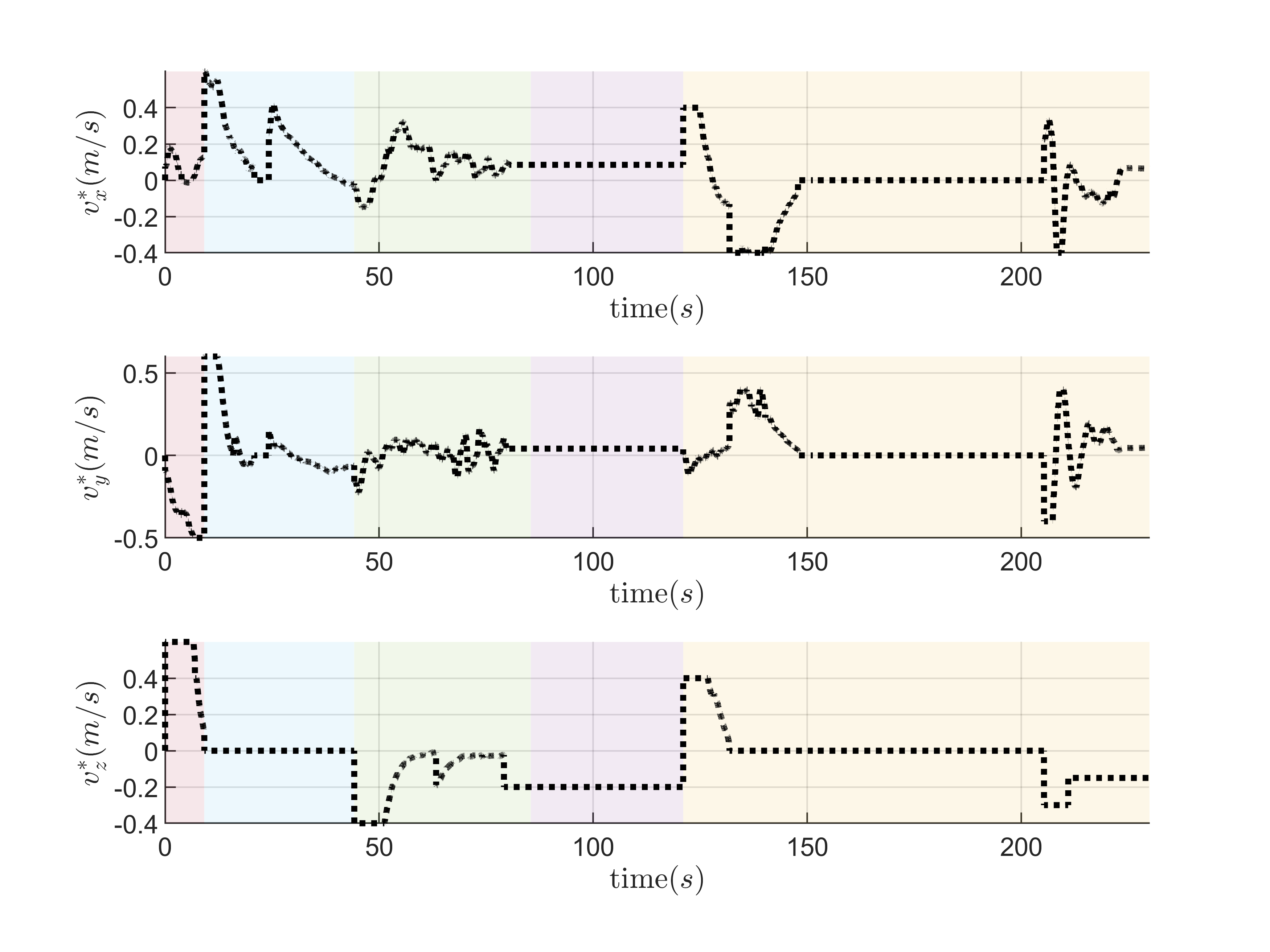}
	\caption{The three-axis velocity command curves of the UAV2 in each process of transportation.}
	\label{fig:vel_2}
\end{figure}

In addition, the control parameters utilized in the UAV's high-level controller throughout various processes are detailed in Table \ref{tab:pid} where $k_p=k_{p_{\{x,y,z\}}},k_i=k_{i_{\{x,y,z\}}}$ and $k_d=k_{d_{\{x,y,z\}}}$.
{\colc For flight safety, the horizontal velocity command is limited to a maximum of $0.6$m/s, and the vertical velocity command is limited to $0.3$m/s for UAV1 and $0.5$m/s for UAV2.}
To prevent indefinite expansion of integral term, the integral control term accumulates the cumulative error over a 3s period and the anti-windup method is implemented. Control rates are computed at a frequency of $50$Hz.
And the yaw angle control is regulated based on a proportional term.
Based on above parameters setting, the Figures \ref{fig:vel_1} and \ref{fig:vel_2} display the velocity control commands for UAV1 and UAV2, respectively, during transportation. The background of these images carries the same significance as the height curve. 

\begin{table}[!htbp] 
\centering 
\vspace{5pt} 
\caption{The PID parameters in different processes} 
\begin{tabular}{c|c|c|c|c|c} 
\hline
Process & $k_{p}$ & $k_{i}$ & $k_{d}$ & $k_{p_\psi}$ & $k_{i_\psi},k_{d_\psi}$\\
\hline
Take-off & 0.8 & 0 & 0.2 & 0.1 & 0\\
Search & 0.5 & 0 & 0 & 0.1 & 0\\
Land & 0.3 & 0.001 & 0.05 & 0.1 & 0\\
Return & 0.1 & 0.001 & 0 & 0.1 & 0\\
\hline
\end{tabular}
\label{tab:pid} 
\end{table}

During the take-off phase, the presence of structural obstacles on the landing platform necessitates an increase in the proportional gain $k_p$ to swiftly navigate the UAV out of the risk zone. Additionally, the differential term coefficient $k_d$ is heightened to reduce oscillation and overshoot.
In the search phase, where high accuracy is not imperative, only proportional control is employed. However, a switching condition must be established to adjust the path points, set to $0.2$m during the competition.
During the landing phase, precision is paramount, prompting the introduction of the integral term $k_i$ to address potential static differences arising from external interference.
In the return phase, changes in the UAV's center of gravity due to the attached cargo may induce oscillations. To mitigate this, the response speed of the aircraft is sacrificed by reducing the proportional control coefficient $k_p$.

\begin{figure}[thpb]
	\centering
    \includegraphics[width=0.8\linewidth]{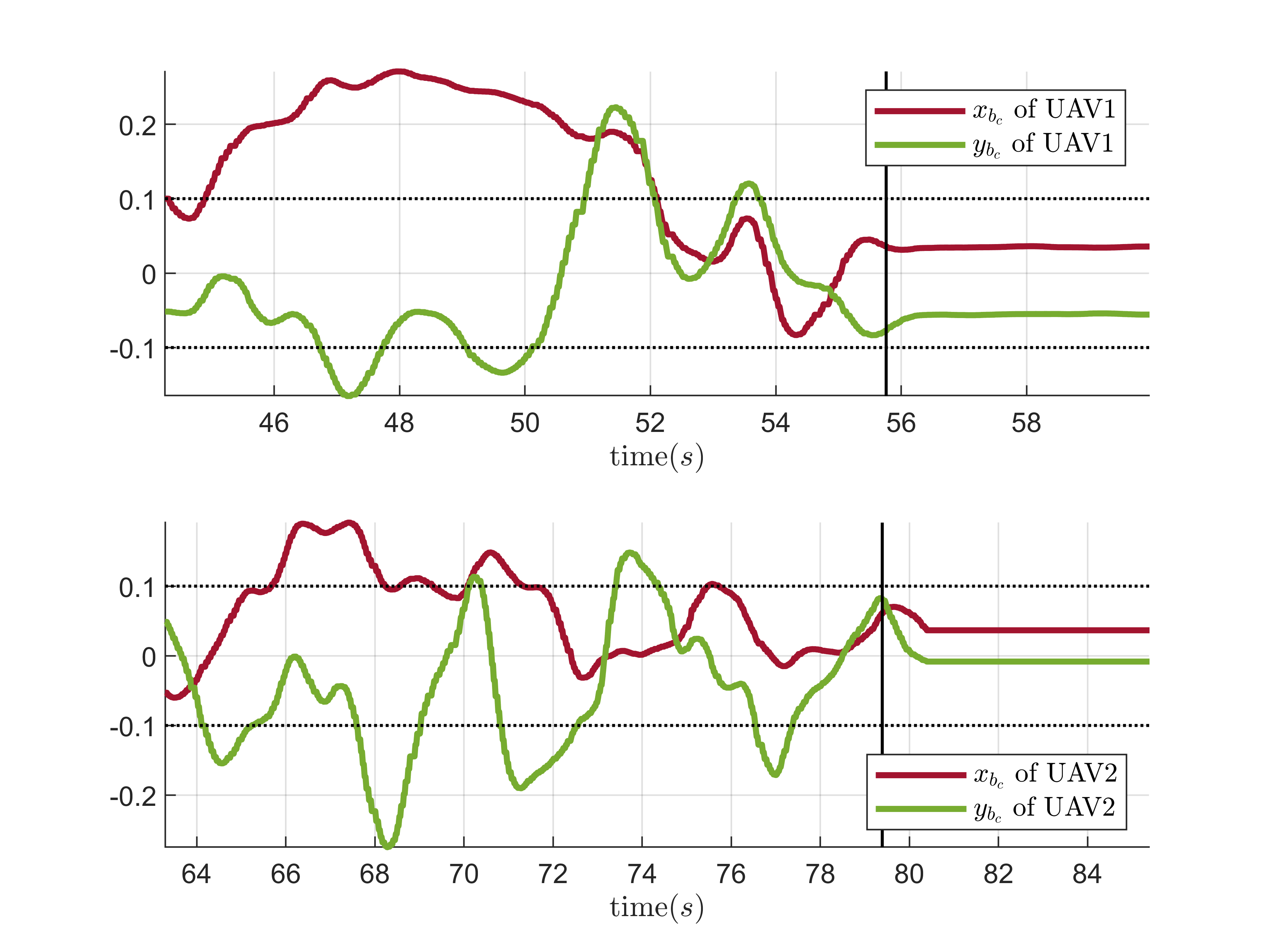}
	\caption{The position estimation curves of the cargo under the body frame in the end of landing phase.}
	\label{fig:land_error}
\end{figure}

{\colb The actual landing position of the UAV determines the contact area between the waterproof adhesive and the cargo, which significantly impacts the success of cargo attachment. To illustrate the landing accuracy for the final competition, we plotted the estimated position curves for UAV1 and UAV2 as they descend below a height of 1m in Figure \ref{fig:land_error}. Due to the absence of true feedback, we treat the estimated position as the horizontal landing error, ignoring estimation error. The red and green curves in the figure represent the estimated position of the cargo along the ${\bf x}_b-$axis and ${\bf y}_b-$axis, respectively. The black dashed line denotes the threshold for blind landing: once the horizontal error falls below this threshold and remains stable for two seconds, the blind landing is executed, marked by the vertical black solid lines. It should be noted that the estimated position to the right of the black solid line is less accurate, as the cargo fills the field of view, making it difficult to extract its true geometric center.}

Based on the results, both UAVs successfully completed their transport tasks despite challenging environmental conditions, yet there remains room for improvement in the system's performance.
Due to time constraints, the transport system currently adopts several conservative strategies, which can be optimized future works.
For instance, further optimization of parameters such as velocity and adsorption time could enhance transportation efficiency.
Integrating multiple sensor data within the UAV is essential to bolster system reliability, while also addressing real-world concerns such as obstacle avoidance. Additionally, streamlining the adsorption structure architecture of the UAV to reduce weight could augment the platform's load capacity.

\section{Conclusion}
This paper presents an innovative approach to autonomous aerial systems designed for precise target localization and transport in GNSS-denied marine environments. The study introduces a comprehensive pipeline for cooperative transport, leveraging both an aerial platform and a non-stationary landing platform.
{\colm The proposed hybrid localization method, based on QR codes and UWB technology, deploys anchors on the landing platform and corrects UAV heading drift. This approach decouples the attitude interactions between the UAV and the landing platform, thereby reducing UAV control oscillations caused by platform motion. While designed for marine environments, this method can also be extended to UAV landing scenarios on ground-based mobile platforms.}
Furthermore, a vision-based servo landing control method, based on a fixed-link camera, is introduced to improve landing stability. Additionally, an innovative grasping mechanism, utilizing a low-cost {\colc adhesive}, is developed to increase the likelihood of successful object capture, even in the presence of wave-induced movements.

{\colc The experimentations without intervention} in the MBZIRC2024 competition validate the robustness, which demonstrates its practical applicability and effectiveness in complex marine environments.
{\colm However, it should be acknowledged that this system was specifically developed for the competition's unique scenario, and significant further research would be needed to generalize it for broader UAV autonomous transport applications. Regarding localization, the current method performs a simple switching between the two localization systems; however, fully integrating visual and distance information would enhance the robustness and accuracy of the system. In more complex scenarios, UAV obstacle avoidance would also need to be further considered, requiring additional sensor data integration and planing algorithms. For system integration, our method design remains limited by access restrictions on certain commercial platforms, such as visual obstacle avoidance sensors. Additionally, this system is primarily designed for single-UAV, single-target transport tasks. Expanding it to multi-UAV collaborative transport holds higher research value and potential applications.}

\subsubsection*{Acknowledgments}
This work is supported by Beijing Institute of Technology under National Key Research and Development Program Grant No. 2022YFE0204400, and the National Natural Science Foundation of China under Grant No. 52302449, and by the Khalifa University under Award No. RC1-2018-KUCARS-8474000136, CIRA-2021-085, MBZIRC-8434000194, KU-BIT-Joint-Lab-8434000534.
\bibliographystyle{apalike}
\bibliography{jfrExampleRefs}

\begin{thebibliography}{}

\bibitem[Anastasiou et~al., 2021]{ref:AHI2021}
Anastasiou, A., Makrigiorgis, R., Kolios, P., and Panayiotou, C. (2021).
\newblock Hyperion: A robust drone-based target tracking system.
\newblock In {\em 2021 International Conference on Unmanned Aircraft Systems (ICUAS)}, pages 927--933.

\bibitem[Andrade et~al., 2022]{ref:FUM2022}
Andrade, F. A.~A., Guedes, I.~P., Carvalho, G.~F., Zachi, A. R.~L., Haddad, D.~B., Almeida, L.~F., de~Melo, A.~G., and Pinto, M.~F. (2022).
\newblock Unmanned aerial vehicles motion control with fuzzy tuning of cascaded-pid gains.
\newblock {\em Machines}, 10(1).

\bibitem[Benini et~al., 2013]{ref:AAJ2013}
Benini, A., Mancini, A., and Longhi, S. (2013).
\newblock An {IMU}/{UWB}/{Vision}-based {Extended} {Kalman} {Filter} for {Mini}-{UAV} {Localization} in {Indoor} {Environment} using 802.15.4a {Wireless} {Sensor} {Network}.
\newblock {\em Journal of Intelligent \& Robotic Systems}, 70(1-4):461--476.

\bibitem[Bernard et~al., 2011]{ref:MAJ2011}
Bernard, M., Kondak, K., Maza, I., and Ollero, A. (2011).
\newblock Autonomous {Transportation} and {Deployment} with {Aerial} {Robots} for {Search} and {Rescue} {Missions}.
\newblock {\em Journal of Field Robotics}, 28(6):914--931.

\bibitem[Bonnin-Pascual and Ortiz, 2019]{ref:BOO2019}
Bonnin-Pascual, F. and Ortiz, A. (2019).
\newblock On the use of robots and vision technologies for the inspection of vessels: A survey on recent advances.
\newblock {\em Ocean Engineering}, 190:106420.

\bibitem[Budiyono and Iwata, 2023]{ref:AAJ2023}
Budiyono, A. and Iwata, K. (2023).
\newblock A survey of the state-of-the-art technologies and innovations in uavs for cargo transportation.
\newblock {\em Journal of Instrumentation, Automation and Systems}, 10(1).

\bibitem[Caballero et~al., 2009]{ref:CVJ2009}
Caballero, F., Merino, L., Ferruz, J., and Ollero, A. (2009).
\newblock Vision-based odometry and slam for medium and high altitude flying uavs.
\newblock {\em Journal of Intelligent and Robotic Systems}, 54:137--161.

\bibitem[Cabreira et~al., 2019]{ref:TSD2019}
Cabreira, T., Brisolara, L., and Ferreira~Jr., P.~R. (2019).
\newblock Survey on {Coverage} {Path} {Planning} with {Unmanned} {Aerial} {Vehicles}.
\newblock {\em Drones}, 3(1):4.

\bibitem[Cadena et~al., 2016]{ref:CPI2016}
Cadena, C., Carlone, L., Carrillo, H., Latif, Y., Scaramuzza, D., Neira, J., Reid, I., and Leonard, J.~J. (2016).
\newblock Past, present, and future of simultaneous localization and mapping: Toward the robust-perception age.
\newblock {\em IEEE Transactions on robotics}, 32(6):1309--1332.

\bibitem[Carvalho et~al., 2023]{ref:FSI2023}
Carvalho, G.~F., Andrade, F. A.~A., Ramos, G.~S., Zachi, A. R.~L., De~Barros, A. L.~F., and Pinto, M.~F. (2023).
\newblock Sliding {Mode} {Controller} {Applied} to {Autonomous} {UAV} {Operation} in {Marine} {Small} {Cargo} {Transport}.
\newblock {\em IEEE Journal on Miniaturization for Air and Space Systems}, 4(4):345--357.

\bibitem[Cassioli et~al., 2002]{ref:DTI2002}
Cassioli, D., Win, M., and Molisch, A. (2002).
\newblock The ultra-wide bandwidth indoor channel: from statistical model to simulations.
\newblock {\em IEEE Journal on Selected Areas in Communications}, 20(6):1247--1257.

\bibitem[Cheng et~al., 2022]{ref:CAJ2022}
Cheng, C., Li, X., Xie, L., and Li, L. (2022).
\newblock Autonomous {Dynamic} {Docking} of {UAV} {Based} on {UWB}-{Vision} in {GPS}-{Denied} {Environment}.
\newblock {\em Journal of the Franklin Institute}, 359(7):2788--2809.

\bibitem[Cho et~al., 2024]{ref:CAK2024}
Cho, J., Jeong, S., and Lee, B. (2024).
\newblock A study on anchor placement and 3d positioning algorithm for uwb application in small sites.
\newblock {\em KSCE Journal of Civil Engineering}, pages 1--13.

\bibitem[Daniel et~al., 2008]{ref:LMI2008}
Daniel, L., Gashinova, M., and Cherniakov, M. (2008).
\newblock Maritime uwb forward scattering radar network: Initial study.
\newblock In {\em 2008 International Conference on Radar}, pages 658--663.

\bibitem[Di~Franco and Buttazzo, 2016]{ref:DJC2016}
Di~Franco, C. and Buttazzo, G. (2016).
\newblock Coverage path planning for uavs photogrammetry with energy and resolution constraints.
\newblock {\em Journal of Intelligent \& Robotic Systems}, 83:445--462.

\bibitem[Ding and Lu, 2021]{ref:CAI2021}
Ding, C. and Lu, L. (2021).
\newblock A {Tilting}-{Rotor} {Unmanned} {Aerial} {Vehicle} for {Enhanced} {Aerial} {Locomotion} and {Manipulation} {Capabilities}: {Design}, {Control}, and {Applications}.
\newblock {\em IEEE/ASME Transactions on Mechatronics}, 26(4):2237--2248.

\bibitem[Feng et~al., 2020]{ref:DKI2020}
Feng, D., Wang, C., He, C., Zhuang, Y., and Xia, X.-G. (2020).
\newblock Kalman-{Filter}-{Based} {Integration} of {IMU} and {UWB} for {High}-{Accuracy} {Indoor} {Positioning} and {Navigation}.
\newblock {\em IEEE Internet of Things Journal}, 7(4):3133--3146.

\bibitem[Ferrigno et~al., 2021]{ref:FAI2021}
Ferrigno, L., Miele, G., Milano, F., Pingerna, V., Cerro, G., and Laracca, M. (2021).
\newblock A uwb-based localization system: analysis of the effect of anchor positions and robustness enhancement in indoor environments.
\newblock In {\em 2021 IEEE International Instrumentation and Measurement Technology Conference (I2MTC)}, pages 1--6.

\bibitem[Gong et~al., 2021]{ref:BRA2021}
Gong, B., Wang, S., Hao, M., Guan, X., and Li, S. (2021).
\newblock Range-based collaborative relative navigation for multiple unmanned aerial vehicles using consensus extended kalman filter.
\newblock {\em Aerospace Science and Technology}, 112:106647.

\bibitem[Grassberger et~al., 1993]{ref:POF1993}
Grassberger, P., Hegger, R., Kantz, H., Schaffrath, C., and Schreiber, T. (1993).
\newblock On {Noise} {Reduction} {Methods} for {Chaotic} {Data}.
\newblock {\em Chaos: An Interdisciplinary Journal of Nonlinear Science}, 3(2):127--141.

\bibitem[Guénot et~al., 2021]{ref:PHA2021}
Guénot, P., Coudreuse, M., Lely, L., and Granger-Veyron, N. (2021).
\newblock Helicopter rescue missions for emergency medical aid at sea: A new assignment for the french military medical service?
\newblock {\em Air Medical Journal}, 40(4):225--231.

\bibitem[Hamel and Samson, 2017]{ref:TPA2017}
Hamel, T. and Samson, C. (2017).
\newblock Position {Estimation} from {Direction} or {Range} {Measurements}.
\newblock {\em Automatica}, 82:137--144.

\bibitem[Hartley and Zisserman, 2003]{ref:RMC2003}
Hartley, R. and Zisserman, A. (2003).
\newblock {\em Multiple view geometry in computer vision}.
\newblock Cambridge university press.

\bibitem[Heredia et~al., 2014]{ref:GCI2014}
Heredia, G., Jimenez-Cano, A., Sanchez, I., Llorente, D., Vega, V., Braga, J., Acosta, J., and Ollero, A. (2014).
\newblock Control of a multirotor outdoor aerial manipulator.
\newblock In {\em 2014 IEEE/RSJ International Conference on Intelligent Robots and Systems}, pages 3417--3422.

\bibitem[Hirano et~al., 2020]{ref:HUI2020}
Hirano, D., Tanishima, N., Bylard, A., and Chen, T.~G. (2020).
\newblock Underactuated gecko adhesive gripper for simple and versatile grasp.
\newblock In {\em 2020 IEEE International Conference on Robotics and Automation (ICRA)}, pages 8964--8969.

\bibitem[Jiang et~al., 2022]{ref:JAP2022}
Jiang, P., Ergu, D., Liu, F., Cai, Y., and Ma, B. (2022).
\newblock A review of yolo algorithm developments.
\newblock {\em Procedia Computer Science}, 199:1066--1073.

\bibitem[Kang et~al., 2022]{ref:KEI2022}
Kang, T.-W., Choi, Y.-S., and Jung, J.-W. (2022).
\newblock Estimation of relative position of drone using fixed size qr code.
\newblock In {\em 2022 IEEE International Conference on Systems, Man, and Cybernetics (SMC)}, pages 442--447.

\bibitem[Kang and Jung, 2024]{ref:TAD2024}
Kang, T.-W. and Jung, J.-W. (2024).
\newblock A {Drone}'s {3D} {Localization} and {Load} {Mapping} {Based} on {QR} {Codes} for {Load} {Management}.
\newblock {\em Drones}, 8(4):130.

\bibitem[Kayton and Fried, 1997]{ref:MAJ1997}
Kayton, M. and Fried, W.~R. (1997).
\newblock {\em Avionics navigation systems}.
\newblock John Wiley \& Sons.

\bibitem[Kudryavtsev et~al., 2018]{ref:AEI2018}
Kudryavtsev, A.~V., Chikhaoui, M.~T., Liadov, A., Rougeot, P., Spindler, F., Rabenorosoa, K., Burgner-Kahrs, J., Tamadazte, B., and Andreff, N. (2018).
\newblock Eye-in-hand visual servoing of concentric tube robots.
\newblock {\em IEEE Robotics and Automation Letters}, 3(3):2315--2321.

\bibitem[Kumar and Behera, 2024]{ref:ADI2024}
Kumar, A. and Behera, L. (2024).
\newblock Design, {Localization}, {Perception}, and {Control} for {GPS}-{Denied} {Autonomous} {Aerial} {Grasping} and {Harvesting}.
\newblock {\em IEEE Robotics and Automation Letters}, 9(4):3538--3545.

\bibitem[Lee et~al., 2010]{ref:TGI2010}
Lee, T., Leok, M., and McClamroch, N.~H. (2010).
\newblock Geometric tracking control of a quadrotor {UAV} on {SE}(3).
\newblock In {\em 49th {IEEE} {Conference} on {Decision} and {Control} ({CDC})}, pages 5420--5425, Atlanta, GA. IEEE.

\bibitem[Li et~al., 2022]{ref:LAI2022}
Li, S., Qiao, L., Zhang, Y., and Yan, J. (2022).
\newblock An early forest fire detection system based on dji m300 drone and h20t camera.
\newblock In {\em 2022 International Conference on Unmanned Aircraft Systems (ICUAS)}, pages 932--937.

\bibitem[Lin and Zhang, 2020]{ref:JLI2020}
Lin, J. and Zhang, F. (2020).
\newblock Loam livox: A fast, robust, high-precision lidar odometry and mapping package for lidars of small fov.
\newblock In {\em 2020 IEEE International Conference on Robotics and Automation (ICRA)}, pages 3126--3131. IEEE.

\bibitem[Liu et~al., 2020a]{ref:LAI2020}
Liu, S., Dong, W., Ma, Z., and Sheng, X. (2020a).
\newblock Adaptive aerial grasping and perching with dual elasticity combined suction cup.
\newblock {\em IEEE Robotics and Automation Letters}, 5(3):4766--4773.

\bibitem[Liu et~al., 2020b]{ref:LRA2020}
Liu, X., Yang, Y., Ma, C., Li, J., and Zhang, S. (2020b).
\newblock Real-time visual tracking of moving targets using a low-cost unmanned aerial vehicle with a 3-axis stabilized gimbal system.
\newblock {\em Applied Sciences}, 10(15):5064.

\bibitem[Mahony et~al., 2012]{ref:MMI2012}
Mahony, R., Kumar, V., and Corke, P. (2012).
\newblock Multirotor aerial vehicles: Modeling, estimation, and control of quadrotor.
\newblock {\em IEEE Robotics \& Automation Magazine}, 19(3):20--32.

\bibitem[Marano et~al., 2010]{ref:SMI2010}
Marano, S., Gifford, W.~M., Wymeersch, H., and Win, M.~Z. (2010).
\newblock Nlos identification and mitigation for localization based on uwb experimental data.
\newblock {\em IEEE Journal on Selected Areas in Communications}, 28(7):1026--1035.

\bibitem[Mellinger et~al., 2011]{ref:DDI2011}
Mellinger, D., Lindsey, Q., Shomin, M., and Kumar, V. (2011).
\newblock Design, modeling, estimation and control for aerial grasping and manipulation.
\newblock In {\em 2011 IEEE/RSJ International Conference on Intelligent Robots and Systems}, pages 2668--2673.

\bibitem[Nowakowski and Idzkowski, 2020]{ref:NUI2020}
Nowakowski, M. and Idzkowski, A. (2020).
\newblock Ultra-wideband signal transmission according to european regulations and typical pulses.
\newblock In {\em 2020 International Conference Mechatronic Systems and Materials (MSM)}, pages 1--4.

\bibitem[Perez-Grau et~al., 2018]{ref:FAJ2018}
Perez-Grau, F.~J., Ragel, R., Caballero, F., Viguria, A., and Ollero, A. (2018).
\newblock An architecture for robust uav navigation in gps-denied areas.
\newblock {\em Journal of Field Robotics}, 35(1):121--145.

\bibitem[Romero et~al., 2007]{ref:RAT2007}
Romero, M., Sheremetov, L., and Soriano, A. (2007).
\newblock {\em A Genetic Algorithm for the Pickup and Delivery Problem: An Application to the Helicopter Offshore Transportation}, pages 435--444.
\newblock Springer Berlin Heidelberg, Berlin, Heidelberg.

\bibitem[Ruan et~al., 2022]{ref:RCI2022}
Ruan, L., Li, G., Dai, W., Tian, S., Fan, G., Wang, J., and Dai, X. (2022).
\newblock Cooperative relative localization for uav swarm in gnss-denied environment: A coalition formation game approach.
\newblock {\em IEEE Internet of Things Journal}, 9(13):11560--11577.

\bibitem[Ruotolo et~al., 2021]{ref:WFC2021}
Ruotolo, W., Brouwer, D., and Cutkosky, M.~R. (2021).
\newblock From grasping to manipulation with gecko-inspired adhesives on a multifinger gripper.
\newblock {\em Science Robotics}, 6(61):eabi9773.

\bibitem[Sang et~al., 2023]{ref:SBI2023}
Sang, C.~L., Adams, M., Hesse, M., and Rückert, U. (2023).
\newblock Bidirectional uwb localization: A review on an elastic positioning scheme for gnss-deprived zones.
\newblock {\em IEEE Journal of Indoor and Seamless Positioning and Navigation}, 1:161--179.

\bibitem[Suarez et~al., 2020]{ref:ABI2020}
Suarez, A., Vega, V.~M., Fernandez, M., Heredia, G., and Ollero, A. (2020).
\newblock Benchmarks for {Aerial} {Manipulation}.
\newblock {\em IEEE Robotics and Automation Letters}, 5(2):2650--2657.

\bibitem[Tahir, 2023]{ref:TFD2023}
Tahir, A. (2023).
\newblock {\em Formation control of swarms of unmanned aerial vehicles}.
\newblock PhD thesis, Doctoral Dissertation, University of Turku, Turku, Finland.

\bibitem[Tahir et~al., 2019]{ref:ASJ2019}
Tahir, A., Boling, J., Haghbayan, M.-H., Toivonen, H.~T., and Plosila, J. (2019).
\newblock Swarms of unmanned aerial vehicles - a survey.
\newblock {\em Journal of Industrial Information Integration}, 16:100106.

\bibitem[Tahir et~al., 2023]{ref:TEI2023}
Tahir, A., Haghbayan, H., Boling, J.~M., and Plosila, J. (2023).
\newblock Energy-efficient post-failure reconfiguration of swarms of unmanned aerial vehicles.
\newblock {\em IEEE Access}, 11:24768--24779.

\bibitem[Thomas and Ros, 2005]{ref:FRI2004}
Thomas, F. and Ros, L. (2005).
\newblock Revisiting {Trilateration} for {Robot} {Localization}.
\newblock {\em IEEE Transactions on Robotics}, 21(1):93--101.

\bibitem[Tong et~al., 2020]{ref:KRI2020}
Tong, K., Wu, Y., and Zhou, F. (2020).
\newblock Recent advances in small object detection based on deep learning: A review.
\newblock {\em Image and Vision Computing}, 97:103910.

\bibitem[Villa et~al., 2020]{ref:DAJ2020}
Villa, D. K.~D., BrandÃ£o, A.~S., and Sarcinelli-Filho, M. (2020).
\newblock A {Survey} on {Load} {Transportation} {Using} {Multirotor} {UAVs}.
\newblock {\em Journal of Intelligent \& Robotic Systems}, 98(2):267--296.

\bibitem[Welch et~al., 1995]{ref:FAC1995}
Welch, G., Bishop, G., et~al. (1995).
\newblock An introduction to the kalman filter.

\bibitem[Wu et~al., 2019]{ref:WSI2019}
Wu, C.-L., Lin, C.-Y., Hirunsirisombut, P., Ng, H.-F., and Shih, T.~K. (2019).
\newblock Searching roi for object detection based on cnn.
\newblock In {\em 2019 International Symposium on Intelligent Signal Processing and Communication Systems (ISPACS)}, pages 1--2.

\bibitem[Yang et~al., 2020]{ref:YMI2020}
Yang, T., Jiang, Z., Sun, R., Cheng, N., and Feng, H. (2020).
\newblock Maritime search and rescue based on group mobile computing for unmanned aerial vehicles and unmanned surface vehicles.
\newblock {\em IEEE Transactions on Industrial Informatics}, 16(12):7700--7708.

\bibitem[Zhang et~al., 2015]{ref:HLI2015}
Zhang, H., Zhang, C., Yang, W., and Chen, C.-Y. (2015).
\newblock Localization and {Navigation} {Using} {QR} {Code} for {Mobile} {Robot} in {Indoor} {Environment}.
\newblock In {\em 2015 {IEEE} {International} {Conference} on {Robotics} and {Biomimetics} ({ROBIO})}, pages 2501--2506, Zhuhai. IEEE.

\bibitem[Zhang et~al., 2024]{ref:ZAI2024}
Zhang, Q., Wang, L., Meng, H., Zhang, W., and Huang, G. (2024).
\newblock A lidar point clouds dataset of ships in a maritime environment.
\newblock {\em IEEE/CAA Journal of Automatica Sinica}, 11(7):1681--1694.

\bibitem[Zhao et~al., 2022]{ref:ZDI2022}
Zhao, M., Li, J., and Wang, H. (2022).
\newblock Detection and recognition of deformed multiple qr codes based on sr\_esagan algorithm.
\newblock In {\em 2022 4th International Conference on Frontiers Technology of Information and Computer (ICFTIC)}, pages 690--696.

\bibitem[Zheng et~al., 2022]{ref:ZLI2022}
Zheng, Z., Yu, Y., Chen, R., Huang, H., Zhao, H., and Lu, X. (2022).
\newblock Localization method based on multi-qr codes for mobile robots.
\newblock In {\em 2022 IEEE International Conference on Advances in Electrical Engineering and Computer Applications (AEECA)}, pages 1385--1391.

\bibitem[Zuo et~al., 2020]{ref:XLI2020}
Zuo, X., Yang, Y., Geneva, P., Lv, J., Liu, Y., Huang, G., and Pollefeys, M. (2020).
\newblock Lic-fusion 2.0: Lidar-inertial-camera odometry with sliding-window plane-feature tracking.
\newblock In {\em 2020 IEEE/RSJ International Conference on Intelligent Robots and Systems (IROS)}, pages 5112--5119. IEEE.

\end{thebibliography}

\end{document}